\pdfoutput=1

\documentclass[11pt]{article}

\PassOptionsToPackage{table}{xcolor}
\usepackage{acl}

\usepackage{times}
\usepackage{latexsym}
\usepackage[T1]{fontenc}
\usepackage[utf8]{inputenc}
\usepackage{microtype}
\usepackage{inconsolata}

\usepackage{enumitem}

\setlist[itemize]{left=0pt}

\usepackage{soul}
\usepackage{url}
\usepackage{pifont}

\usepackage{graphicx}
\usepackage{subcaption}
\usepackage{stfloats}
\usepackage{float}

\usepackage{booktabs}
\usepackage{multirow}
\usepackage{tabularx}
\usepackage{longtable}
\usepackage{rotating}
\usepackage{array}
\usepackage{seqsplit}

\usepackage{amsmath}
\usepackage{amsthm}
\usepackage{amssymb}

\usepackage{algorithm}
\usepackage{algorithmic}

\title{
SUP-MIMIC: A Multi-Task Clinical Diagnosis Benchmark for Evaluating LLMs' Robustness to Contradictory Evidence
}

\author{
Yu Yi\textsuperscript{1,*}
\quad
Wang Bo\textsuperscript{2,*}
\quad
Feng Chong\textsuperscript{2}
\quad
Shi Ge\textsuperscript{2,\textdagger}
\\
Liu Xia\textsuperscript{3}
\quad
Yang Ziyi\textsuperscript{1}
\quad
Ye Xiang\textsuperscript{1}
\quad
Shi Xuewen\textsuperscript{4}
\\[0.4em]
\textsuperscript{1}Beijing University of Technology
\quad
\textsuperscript{2}Beijing Institute of Technology
\\
\textsuperscript{3}Department of Rheumatology and Immunology,
China-Japan Friendship Hospital
\\
\textsuperscript{4}Dongbei University of Finance and Economics
\\[0.3em]
\textsuperscript{*}Equal contribution.
\qquad
\textsuperscript{\textdagger}Corresponding author.
}

\begin{document}

\maketitle

% =========================================================
% Abstract
% =========================================================

\begin{abstract}
%% 背景→问题→方法→结果→意义

Current evaluations of large language models (LLMs) primarily focus on factual knowledge retrieval, overlooking the fundamental challenge of navigating the complex, non-bijective mappings between clinical indicators and diagnoses. Existing benchmarks fail to assess whether large language models truly possess the reasoning capability required for diagnostic ambiguity scenarios, where identical clinical presentations may correspond to different etiologies, and diagnostic convergence scenarios, where heterogeneous symptoms ultimately indicate the same disease.
To address this issue, we propose the SUP-MIMIC, a multi-task framework utilizing MIMIC-IV-v3.1 that comprises Basic Assessment (BA), Diagnostic Divergence Task (DDT), and Diagnostic Convergence Task (DCT).  Specifically, DDT is designed to evaluate the model’s “one-to-many” disambiguation capability among phenotypically similar cases, while DCT assesses the model’s ability to identify “many-to-one” diagnostic patterns across different pathophysiological pathways.
Comprehensive evaluation of state-of-the-art LLMs reveals substantial performance degradation on DDT and DCT compared to baseline tasks, exposing a systemic reliance on statistical shortcuts over genuine causal reasoning. Our findings further highlight a conservative bias toward "healthy" predictions, implying non-trivial risks for missed diagnoses in realistic medical settings. This work establishes a rigorous methodology for quantifying clinical reasoning robustness and provides a roadmap for enhancing the safety of language models in clinical medicine.

\end{abstract}
\section{Introduction}
\begin{figure}[t]
    \centering
    \includegraphics[width=0.48\textwidth]{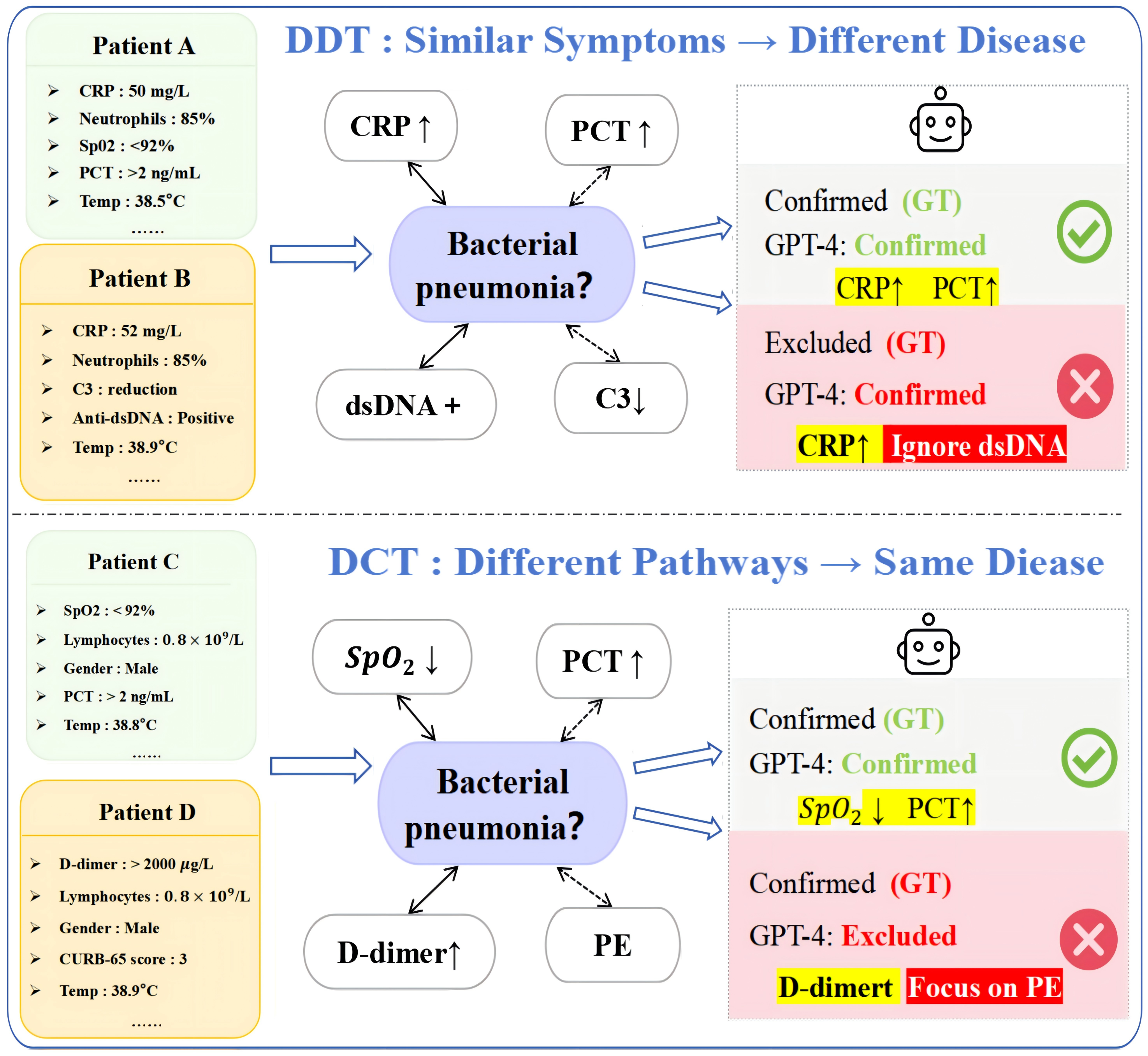}
    \caption{Illustration of two adversarial diagnostic contrasts. In Diagnostic Divergence, Patient~A and Patient~B share similar clinical profiles, yet the model correctly identifies bacterial pneumonia in Patient~A but fails to recognize SLE in Patient~B despite decisive laboratory differences. In Diagnostic Convergence, both Patient~C and Patient~D have bacterial pneumonia, yet the model only recognizes the classic presentation in Patient~C and misses the same diagnosis in Patient~D.}
    \label{fig:intro}
\end{figure}
%% P1: Benchmark Gap
Large language models (LLMs) have demonstrated strong performance on medical question answering and clinical diagnosis benchmarks such as MedQA \cite{jin2020medqa}, PubMedQA \cite{jin2019pubmedqa}, MMLU-Medical \cite{hendrycks2021measuringmassivemultitasklanguage}, and LLMEval-Med \cite{zhang-etal-2025-llmeval}. However, these evaluations predominantly assess medical knowledge recall or isolated single-case diagnosis, providing limited insight into whether models remain reliable when clinical evidence is ambiguous, conflicting, or misleading. Existing benchmarks rarely provide a systematic evaluation of LLM robustness under two pervasive clinical phenomena: \textit{diagnostic divergence}, where clinically similar patients require different diagnoses, and \textit{diagnostic convergence}, where heterogeneous clinical presentations correspond to the same diagnosis.

%% P2: Clinical Motivation
In clinical practice, patient features and diagnostic labels do not follow a simple one-to-one mapping. Patients with similar demographics, laboratory values, and comorbidity profiles may require different diagnoses because a small set of clinical indicators can change the diagnostic interpretation. Conversely, patients with markedly different presentations may still receive the same diagnosis. This feature-label misalignment creates a robustness challenge for LLMs: reliable diagnosis requires sensitivity to decisive differences among similar cases and invariance to heterogeneity that does not alter the diagnosis.

%% P3: Motivating Example (Figure 1)
Figure~\ref{fig:intro} illustrates these two failure modes with contrastive patient pairs. In the Diagnostic Divergence case, Patient~A and Patient~B have highly similar clinical profiles, but only Patient~A has bacterial pneumonia. GPT-4 correctly identifies pneumonia in Patient~A from indicators such as \textit{PCT} and \textit{SpO$_2$}, yet incorrectly assigns a similar diagnosis to Patient~B despite evidence supporting an SLE flare, including \textit{Anti-dsDNA}$^+$ and \textit{C3}$\downarrow$. In the Diagnostic Convergence case, both Patient~C and Patient~D have bacterial pneumonia, but their clinical presentations differ substantially. GPT-4 recognizes the classic presentation in Patient~C while missing the same diagnosis in Patient~D, whose evidence includes \textit{D-dimer} and \textit{CURB-65}. These examples suggest that models relying on surface similarity can fail to maintain diagnostic consistency across heterogeneous presentations.

%% P4: Proposed Benchmark
To address this gap, we introduce \textbf{SUP-MIMIC}, a multi-task clinical diagnosis benchmark designed to evaluate LLM robustness to contradictory evidence. We construct SUP-MIMIC from MIMIC-IV-v3.1 ICU records by applying supervised feature ranking and patient-level similarity estimation to mine naturally occurring adversarial patient pairs from real clinical data. The benchmark contains three evaluation layers. \textbf{Basic Assessment (BA)} evaluates single-case diagnosis across 200 prevalent diseases. \textbf{Diagnostic Divergence Task (DDT)} tests whether models can distinguish different diagnoses among clinically similar patients. \textbf{Diagnostic Convergence Task (DCT)} tests whether models can identify the same diagnosis across clinically heterogeneous presentations.

%% P5: Findings Preview
Our experiments show that models achieve substantially higher accuracy on BA than on DDT and DCT, indicating that conventional single-case evaluation can obscure failures under contrastive clinical evidence. We further introduce pair-aware robustness metrics that assess whether model predictions preserve the diagnostic relationship within each patient pair. These metrics reveal substantial pairwise inconsistency and a pervasive healthy prediction bias, where models over-predict the absence of disease in challenging cases and thereby increase the risk of missed diagnoses.

%% P6: Contributions
Our contributions are as follows:
\begin{itemize}
    \item We introduce \textbf{SUP-MIMIC}, a pairwise adversarial clinical diagnosis benchmark for evaluating LLM robustness to contradictory evidence in ICU diagnosis. We formalize two complementary diagnostic robustness tasks: DDT, which evaluates sensitivity to decisive differences among clinically similar patients, and DCT, which evaluates invariance across heterogeneous presentations of the same diagnosis.
    \item We develop a data-driven adversarial pair mining pipeline that combines clinically informative feature ranking with patient-level similarity estimation to construct challenging diagnosis pairs from real ICU records.
    \item We evaluate thirteen contemporary LLMs with pair-aware robustness metrics, showing that strong single-case accuracy can coexist with substantial pairwise inconsistency and systematic failure modes such as shortcut reliance and healthy prediction bias.
\end{itemize}

\section{Related work}
\paragraph{Medical Diagnostic Benchmark Datasets.}
Medical AI has advanced through high-quality clinical datasets and benchmarks\cite{BLAGEC2023104274,GAO2023104286}. Early datasets like MIMIC-III\cite{Harutyunyan_2019} enabled clinical prediction tasks, supporting disease prediction\cite{PMID:34277655}, patient prognosis\cite{Qian2023}, and personalized treatment\cite{Wang2023}. Standardization efforts (e.g., MIMIC-Extract\cite{10.1145/3368555.3384469}) facilitated cross-model comparisons, while EHRShot\cite{wornow2023ehrshotehrbenchmarkfewshot} validated few-shot transferability. Other datasets (eICU\cite{Pollard2018}, VitalDB\cite{Lee2022}) expanded time-series prediction and risk stratification, though evaluation frameworks remained scenario-specific.

With NLP advancements, focus shifted to clinical text comprehension. EHRNoteQA\cite{Kweon2024} evaluated information extraction, while MedQA\cite{jin2020diseasedoespatienthave} and LLM-MedQA\cite{yang2025llmmedqaenhancingmedicalquestion} assessed medical knowledge via USMLE-style questions. Retrieval-augmented generation (RAG) improved QA accuracy by 18\%\cite{yu2024rankragunifyingcontextranking}. LLMs further spurred diagnostic evaluations through benchmarks like MedMCQA\cite{pmlr-v174-pal22a} and ClinicalGPT\cite{wang2023clinicalgptlargelanguagemodels}.

However, existing benchmarks lack causal reasoning assessment, primarily testing surface-level pattern matching rather than pathophysiological understanding. This prevents validation of models' ability to distinguish spurious correlations from causal relationships. To address this gap, we propose a framework defining diagnostic scenarios for causal reasoning evaluation, with an automated pipeline for adversarial dataset construction.
%C3+C2
%C2: Systematically identifying clinical features and matching sample pairs (the process of constructing the dataset)
%C3: Quantitative Model, Evaluation Model (Evaluation Model)

\paragraph{Medical Causal Discovery and Diagnostic Consistency Research.}
Medical causal discovery methods comprise two categories: statistical association analysis (e.g., temporal inference in MIMIC-IV-Causal\cite{Li2024CISepsis}, imaging correlations in RadCausality-12K\cite{7470527}) and case-based validation (e.g., UK Biobank-Causal\cite{Lehrer2021Endometrial}). Advanced approaches integrate knowledge graphs (ClinicalGPT\cite{wang2023clinicalgptlargelanguagemodels}) and semantic similarity (DDT-LLM\cite{yang2025improvingfactualitylargelanguage}), reducing analysis errors by 22\%. BioLinkBERT enhances causal phrase labeling\cite{yasunaga2022linkbertpretraininglanguagemodels}.

Diagnostic consistency evaluations follow two paradigms: rule-based checks (NCCN\cite{Gradishar2016}, AHA\cite{Bozkurt2021}, WHO standards) and similarity matching (Mayo Clinic\cite{cooper2009revised}, DeepLesion\cite{yan2018deeplesion}). These are constrained by predefined rules or superficial features, failing to capture deeper commonalities.

However, existing methods lack a systematic framework for evaluating deep feature-diagnosis association reasoning in complex medical diagnosis, and merely perform fragmented pattern matching without distinguishing spurious correlations from genuine associative relationships. To fill this gap, we formalize the non-bijective nature of clinical reasoning, build adversarial evaluation datasets to systematically probe LLMs' performance on contradictory diagnostic mappings, and validate their ability to grasp and leverage implicit pathophysiological related relationships.

%%C1: Two tasks (datasets)

\section{Method}

\begin{figure}[htbp]
     \centering
     \includegraphics[width=0.48\textwidth]{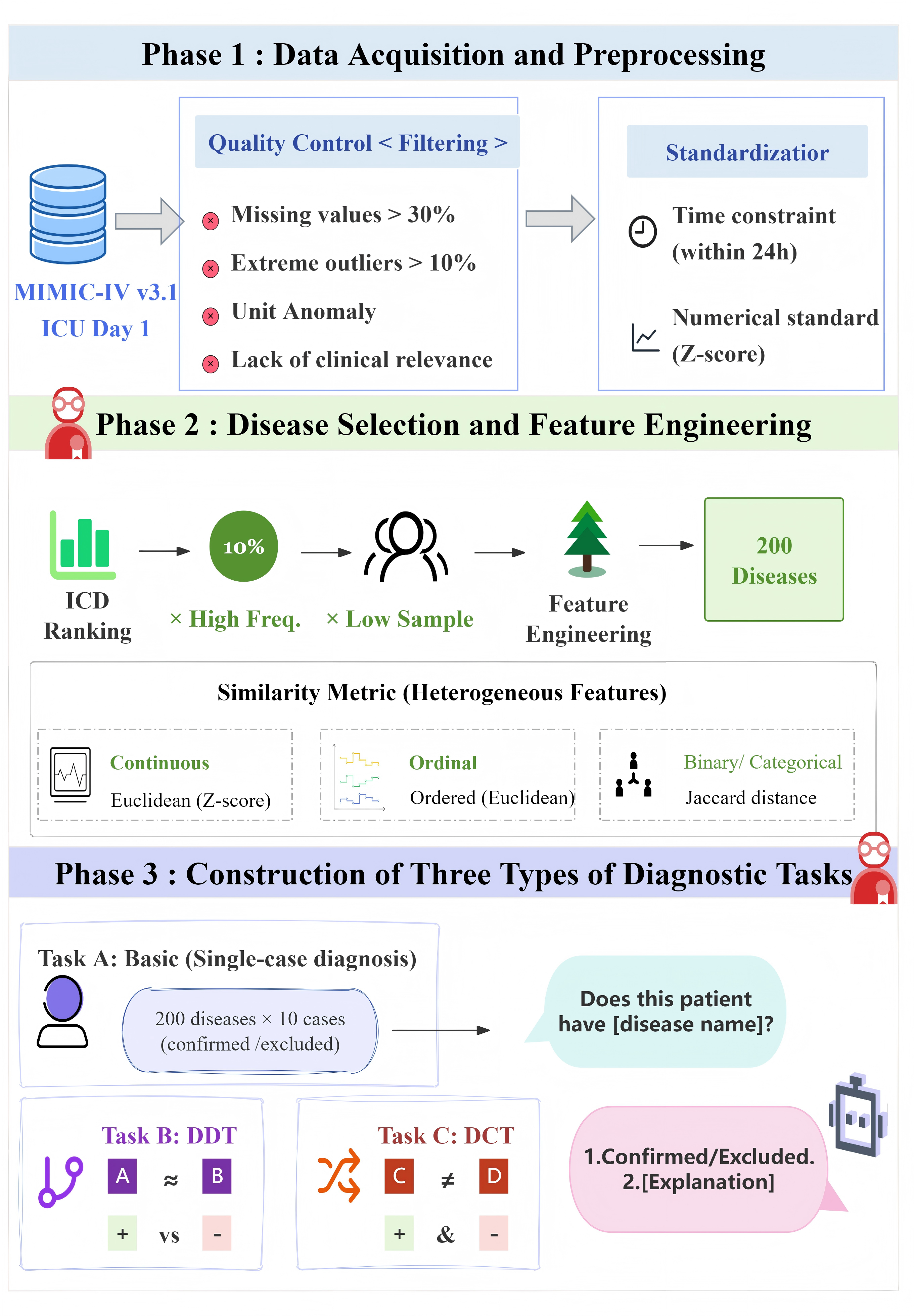}
     \caption{Overview of the SUP-MIMIC construction pipeline. Multi-label ICU records are converted into binary patient-diagnosis verification labels. The full 448-dimensional feature space is used as LLM input, while diagnosis-specific top-$K_f$ features are used for similarity estimation and adversarial pair mining across BA, DDT, and DCT.}
     \label{fig:mth}
\end{figure}

We formulate clinical diagnosis as binary verification over multi-label patient records. Each patient $p$ has clinical indicators $\mathbf{X}_p$ and ICD-coded diagnoses $Y_p\subseteq \mathcal{D}$. Given candidate diagnosis $d_k$, the model predicts $y_{p,k}=\mathbb{1}[d_k\in Y_p]$. The full 448-dimensional feature vector is used for LLM input, while diagnosis-specific top-$K_f$ features support pair mining and binary patient-diagnosis labels support evaluation. This section describes data preprocessing, disease and indicator selection, similarity estimation, and adversarial pair construction, as illustrated in Figure~\ref{fig:mth}.

\subsection{Data Filtering and Preprocessing}

We use the MIMIC-IV-v3.1 dataset\footnote{\url{https://physionet.org/content/mimiciv/3.1/}}~\cite{mimiciii}, focusing on intensive care unit (ICU) records. We restrict clinical measurements to the first 24 hours after ICU admission and use ICD-coded diagnoses associated with the ICU stay as verification labels. This window aligns model inputs with early clinical evidence and reduces reliance on measurements observed later in the stay. We exclude records with incomplete core information or clinically implausible values. Retained features with limited missingness are imputed using mean values or temporal interpolation when repeated measurements are available, and patient records are kept only when core structured variables such as vital signs, laboratory results, and clinical scores are available.

\subsection{Base Task Construction}

We first construct the BA as the single-case diagnostic verification layer of SUP-MIMIC. We rank all ICD-coded diagnoses by frequency and exclude the top 10\% most frequent labels, which tend to be broad categories that overwhelm the benchmark distribution. We also remove diagnoses with fewer than 100 positive cases and select $M=200$ representative diseases with at least 70\% data completeness and no severe outlier patterns.

For each selected diagnosis $d_k \in \mathcal{D}$, we train a Random Forest classifier to  rank clinically informative features for pair mining. We train a Random Forest classifier using patients with $d_k$ as positive examples and an equal number of randomly sampled patients without $d_k$ as negative examples. We then rank clinical indicators by their feature importance scores and select the top $K_f$ indicators as the diagnosis-specific feature set $F_k$. These feature sets are used only for patient similarity estimation and adversarial pair construction. During LLM evaluation, the model receives all clinical features rather than only the top-$K_f$ indicators.

\begin{figure}[htbp]
     \centering
     \includegraphics[width=0.48\textwidth]{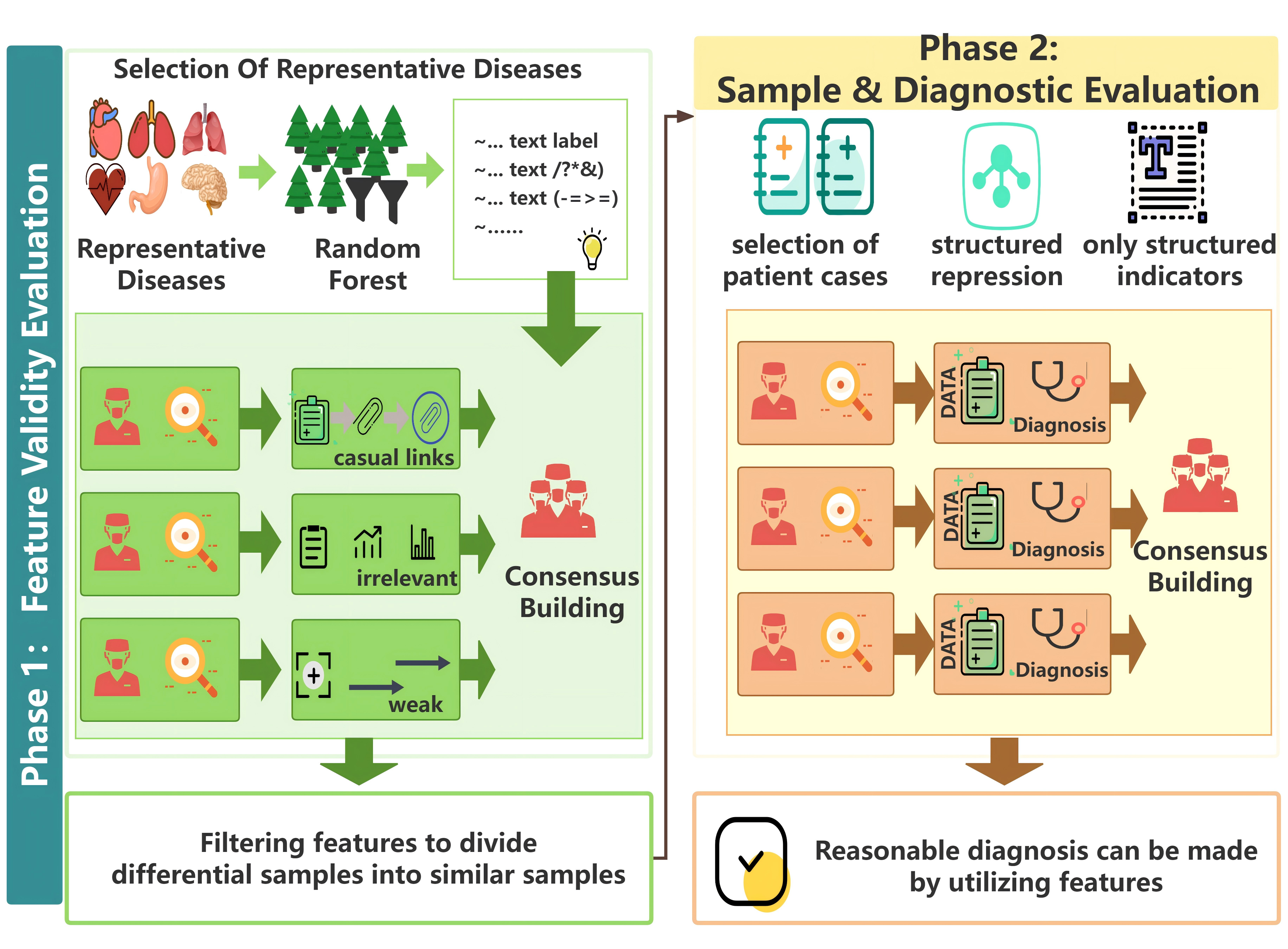}
     \caption{Diagram of the two-stage expert validation.}
     \label{fig:Expert}
\end{figure}

\subsection{Case Similarity and Adversarial Pair Construction}

To construct the Diagnostic Divergence Task (DDT) and Diagnostic Convergence Task (DCT), we compute patient similarity in an anchor-specific feature space. For each diagnosis $d_k$, similarity is measured only over its selected feature set $F_k$.

For patients $p_i$ and $p_j$, we define
\begin{equation}
\mathrm{dist}_k(p_i,p_j)
=
\frac{1}{|F_k|}
\sum_{f\in F_k}
\delta_f(x_{i,f},x_{j,f}),
\end{equation}

where $\delta_f$ is defined according to the feature type. For continuous variables, values are min--max normalized and compared by absolute difference:
\begin{equation}
\delta_f(x_{i,f}, x_{j,f}) = |z(x_{i,f}) - z(x_{j,f})|.
\end{equation}
For ordinal variables, ordered categories are mapped to integer values, normalized, and compared in the same way. For binary or categorical variables, we use an indicator distance:
\begin{equation}
\delta_f(x_{i,f}, x_{j,f}) = \mathbb{1}[x_{i,f} \neq x_{j,f}].
\end{equation}
This per-feature formulation normalizes scale differences and prevents the distance from being dominated by any single variable type.

For each anchor diagnosis $d_k$, we estimate a diagnosis-specific similarity distribution by computing $\mathrm{dist}_k(p_i,p_j)$ for patient pairs satisfying $y_{i,k}=1$ and $y_{j,k}=1$. Let $\theta_k$ and $\phi_k$ denote the 25th and 75th percentiles of this within-diagnosis distance distribution. Patient pairs are considered similar with respect to $d_k$ if $\mathrm{dist}_k(p_i,p_j)<\theta_k$, and highly dissimilar if $\mathrm{dist}_k(p_i,p_j)>\phi_k$.

We then construct two sets of naturally occurring adversarial pairs. Here, ``adversarial'' refers to contrastive patient pairs mined from real clinical records where feature similarity and diagnostic labels are deliberately put in tension; it does not refer to synthetic perturbations or model-specific attacks.

For \textbf{DDT}, we mine hard negative pairs for each anchor diagnosis $d_k$:

\begin{equation}
\mathcal{P}_{\mathrm{DDT}}^k
=
\{(i,j,k): y_{i,k}\!=\!1,\; y_{j,k}\!=\!0,\; d^k_{ij}<\theta_k\}.
\end{equation}

These pairs contain a patient with $d_k$ and a clinically similar patient without $d_k$, testing whether models can reject the anchor diagnosis despite superficial similarity.

For \textbf{DCT}, we mine heterogeneous positive pairs for each anchor diagnosis $d_k$:

\begin{equation}
\mathcal{P}_{\mathrm{DCT}}^k
=
\{(i,j,k): y_{i,k}\!=\!1,\; y_{j,k}\!=\!1,\; d^k_{ij}>\phi_k\}.
\end{equation}

These pairs contain two patients who share the anchor diagnosis but differ substantially in the anchor-specific feature space, testing whether models can recognize the same diagnosis across heterogeneous presentations.

For each diagnosis, we retain the top-$m$ eligible pairs by distance (smallest for DDT, largest for DCT). For a fixed anchor $d_k$, the two sets are disjoint by construction because they impose mutually exclusive labels on $p_j$. $y_{j,k}=0$ for DDT and $y_{j,k}=1$ for DCT. Across anchors, the same patient may appear as a distinct $(p,d_k)$ verification instance, and all distances are computed within the corresponding $F_k$.

\subsection{Expert Validation}

We conduct two-stage expert validation to assess the clinical validity of selected features and adversarial pairs (Figure~\ref{fig:Expert}). In Stage~1, experts independently review disease-specific feature sets for randomly sampled benchmark diseases and categorize each feature as a diagnostic indicator, supportive but nonspecific correlate, weakly informative feature, or unclear/irrelevant feature. In Stage~2, experts review complete structured feature vectors for sampled cases and make diagnostic judgments for the target diagnosis using only the provided indicators, without free-text notes or imaging reports. This protocol assesses whether the benchmark contains clinically interpretable evidence for the intended verification tasks. Expert panel details and agreement statistics are reported in Section~\ref{subsec:quality}.

\section{Experiments}

\subsection{Evaluation Metrics}

We evaluate models on three SUP-MIMIC tasks: Basic Assessment (BA), Diagnostic Divergence Task (DDT), and Diagnostic Convergence Task (DCT). Each instance is formulated as binary diagnostic verification for an anchor diagnosis $d_k$. Given a patient $p$, the model outputs $\hat{y}_{p,k}\in\{0,1\}$, indicating whether $d_k$ is supported by the patient's clinical profile. For all tasks, we report overall accuracy together with Sick Recall and Healthy Recall.

\paragraph{\textbf{Average Diagnostic Robustness (ADR).}}
To summarize performance across standard and adversarial settings, we define ADR as the mean of task-level accuracy and pairwise robustness:
\begin{equation}
\mathrm{ADR}
=
\frac{
\mathrm{Acc}_{\text{BA}}
+
\mathrm{PDRA}_{\text{DDT}}
+
\mathrm{PDRA}_{\text{DCT}}
}{3}.
\end{equation}
ADR provides a simple aggregate view of model performance across the three diagnostic verification settings.

\paragraph{\textbf{Pairwise Diagnostic Robustness Accuracy (PDRA).}}
Because DDT and DCT are pairwise tasks, we evaluate whether a model preserves the correct diagnostic relationship within each adversarial pair. For both tasks, a pair is counted as correct only when the model makes the correct binary verification decision for both patients with respect to the same anchor diagnosis.

For DDT, each pair $(p_i,p_j,d_k)$ contains a positive patient ($y_{i,k}=1$) and a hard negative patient ($y_{j,k}=0$). The pair is correct only if the model accepts $d_k$ for $p_i$ and rejects $d_k$ for $p_j$:
\begin{equation}
\mathrm{PDRA}_{\text{DDT}}
\!=\!
\frac{1}{|\mathcal{P}_{\text{DDT}}|}
\!\sum_{(i,j,k)\in \mathcal{P}_{\text{DDT}}}\!
\mathbb{1}[\hat{y}_{i,k}\!=\!1 \,{\land}\, \hat{y}_{j,k}\!=\!0].
\end{equation}

For DCT, each pair contains two positive patients ($y_{i,k}=1$, $y_{j,k}=1$). The pair is correct only if the model accepts $d_k$ for both:
\begin{equation}
\mathrm{PDRA}_{\text{DCT}}
\!=\!
\frac{1}{|\mathcal{P}_{\text{DCT}}|}
\!\sum_{(i,j,k)\in \mathcal{P}_{\text{DCT}}}\!
\mathbb{1}[\hat{y}_{i,k}\!=\!1 \,{\land}\, \hat{y}_{j,k}\!=\!1].
\end{equation}

\paragraph{\textbf{SUP-MIMIC Robustness Score (SRS).}}
To jointly evaluate standard diagnostic accuracy and adversarial pairwise robustness, we define the SUP-MIMIC Robustness Score:
\begin{equation}
\mathrm{SRS}
=
\mathrm{Acc}_{\text{BA}}\!\cdot\!
\sqrt{\mathrm{PDRA}_{\text{DDT}} \cdot \mathrm{PDRA}_{\text{DCT}}}.
\end{equation}
The geometric mean penalizes models that perform well on one adversarial task but poorly on the other, while $\mathrm{Acc}_{\text{BA}}$ ensures that the composite score remains grounded in basic diagnostic performance. We report SRS as a compact summary and analyze its components separately.

\begin{figure}[t]
    \centering
    \includegraphics[width=\linewidth]{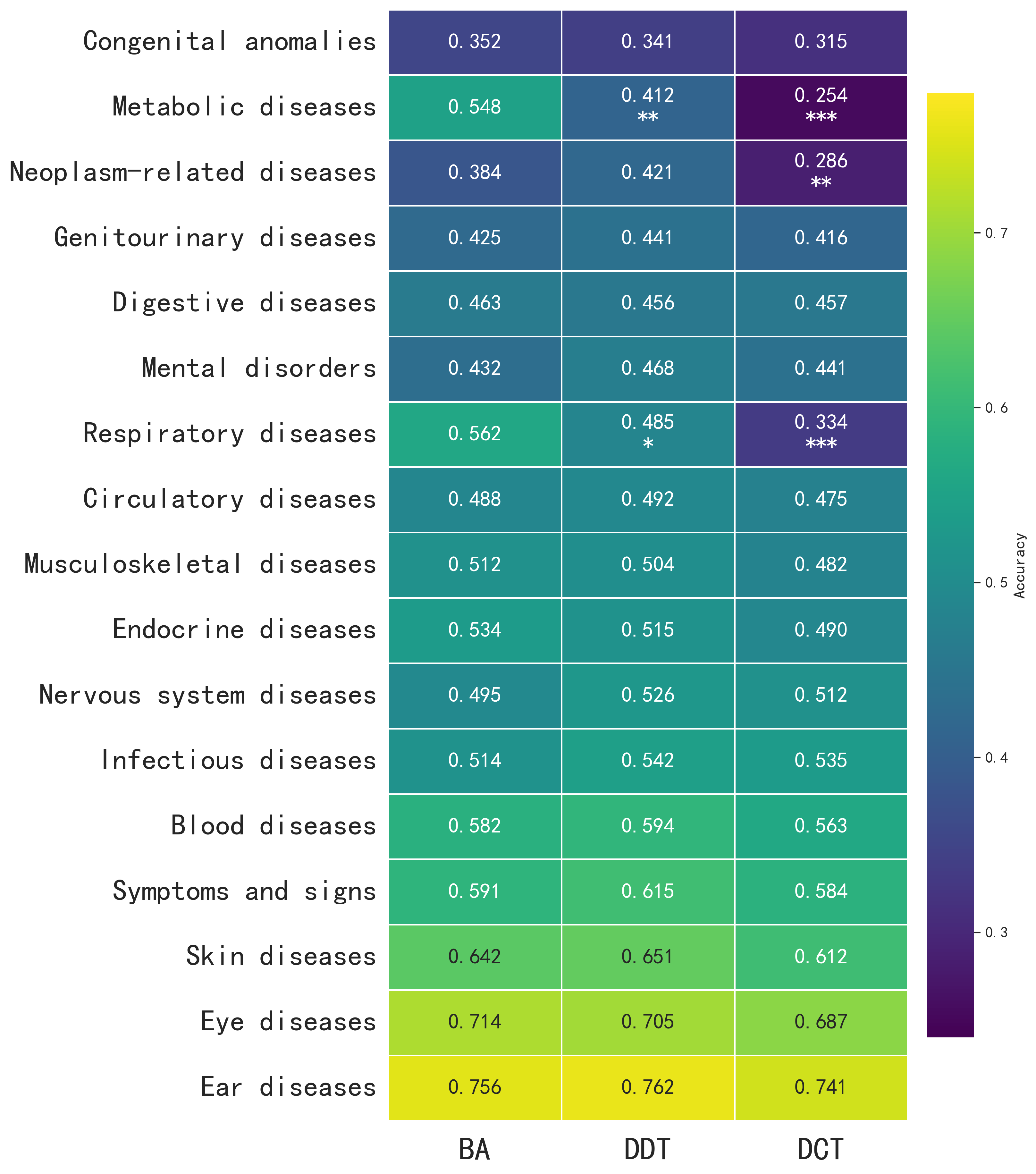}
    \caption{Accuracy heatmap for 17 clinical categories. Categories defined by distinctive single-system biomarker profiles (top rows) maintain stable accuracy, whereas multi-system categories (bottom rows) exhibit significant DCT collapse. Significance markers denote within-category paired $t$-tests against BA ($^{*}p < 0.05$, $^{**}p < 0.01$, $^{***}p < 0.001$).}
    \label{fig:stratification}
\end{figure}

\subsection{Main Results}
\label{subsec:main-results}

\begin{table*}[htbp]
\scriptsize
\centering
\renewcommand{\arraystretch}{1.05}
\resizebox{\textwidth}{!}{%
\begin{tabular}{
    l
    c % Acc_BA
    c % Acc_DDT
    c % Acc_DCT
    c % ADR
    c % PDRA_DDT
    c % PDRA_DCT
    c % SRS
    c % Sick_R
    c % Healthy_R
}
\toprule
\textbf{Model} &
\textbf{$\mathrm{Acc}_{\text{BA}}$} &
\textbf{$\mathrm{Acc}_{\text{DDT}}$} &
\textbf{$\mathrm{Acc}_{\text{DCT}}$} &
\textbf{ADR} &
\textbf{$\mathrm{PDRA}_{\text{DDT}}$} &
\textbf{$\mathrm{PDRA}_{\text{DCT}}$} &
\textbf{SRS} &
\textbf{Sick$_R$} &
\textbf{Healthy$_R$} \\
\midrule
\multicolumn{10}{c}{\cellcolor{gray!20}\textit{Open-Source General LLMs}} \\
\textbf{Llama3.3-70B} &
0.3572 & 0.5806 & 0.3659 & 0.4346 & 0.2038 & 0.2214 & 0.1646 & 0.3638 & 0.7964 \\
\textbf{Qwen2.5-7B} &
0.2986 & 0.4995 & 0.3115 & 0.3699 & 0.2134 & 0.2934 & 0.1178 & 0.3025 & 0.7066 \\
\textbf{Qwen2.5-14B} &
0.3263 & 0.5795 & 0.2825 & 0.3961 & 0.2256 & 0.3026 & 0.1320 & 0.3232 & 0.7606 \\
\textbf{Qwen2.5-32B} &
$\mathbf{0.5619}$ & 0.6163 & 0.3462 & 0.5081 & 0.2694 & 0.3428 & 0.2595 & 0.4450 & 0.8240 \\
\textbf{Mistral-7B} &
0.2639 & 0.4927 & 0.1561 & 0.3042 & 0.3028 & 0.2802 & 0.0732 & 0.2266 & 0.6924 \\
\textbf{DeepSeek-V3} &
0.4769 & 0.6070 & $\mathbf{0.5287}$ & 0.5375 & 0.3050 & 0.2956 & 0.2702 & 0.4796 & $\underline{0.8272}$ \\
\textbf{GLM-4.7} &
0.2564 & $\underline{0.6305}$ & 0.3489 & 0.4119 & 0.2262 & 0.2530 & 0.1203 & 0.3318 & 0.8126 \\
\midrule
\multicolumn{10}{c}{\cellcolor{gray!20}\textit{Open-Source Medical LLMs}} \\
\textbf{HuatuoGPT-o1-8B} &
$\underline{0.5222}$ & 0.6015 & 0.5055 & $\underline{0.5431}$ &
$\underline{0.3342}$ & $\mathbf{0.3976}$ & $\mathbf{0.2879}$ &
$\underline{0.4922}$ & 0.7934 \\
\textbf{MedReason-8B} &
0.4384 & 0.6079 & 0.4655 & 0.5039 & 0.2244 & 0.2894 & 0.2332 & 0.4424 & 0.8116 \\
\midrule
\multicolumn{10}{c}{\cellcolor{gray!20}\textit{Closed-Source General LLMs}} \\
\textbf{GPT-3.5} &
0.4036 & 0.5290 & 0.4421 & 0.4582 & 0.3166 & $\underline{0.3450}$ &
0.1952 & 0.4036 & 0.7316 \\
\textbf{GPT-4o} &
0.4821 & $\mathbf{0.6766}$ & $\underline{0.5172}$ & $\mathbf{0.5586}$ &
$\mathbf{0.3662}$ & 0.2990 & $\underline{0.2852}$ &
$\mathbf{0.4999}$ & $\mathbf{0.8504}$ \\
\textbf{Gemini-2.5 Flash} &
0.3983 & 0.6184 & 0.4833 & 0.5000 & 0.3000 & 0.3236 & 0.2177 & 0.4420 & 0.7898 \\
\textbf{Claude Sonnet 4.5} &
0.2654 & 0.5737 & 0.4061 & 0.4151 & 0.3128 & 0.2820 & 0.1281 & 0.3500 & 0.7478 \\
\bottomrule
\end{tabular}
}
\caption{
Main results on SUP-MIMIC. All metrics are reported as means over five runs; standard deviations are omitted for brevity.
The best result in each column is \textbf{bolded} and the second-best is \underline{underlined}.
}
\label{tab:main-unified}
\end{table*}

Table~\ref{tab:main-unified} reports mean performance across five independent runs. Three principal findings emerge.

\textbf{Pairwise evaluation exposes reasoning inconsistency masked by pointwise accuracy.} Most models achieve higher accuracy on DDT than on BA because each anchor diagnosis is rare in the patient pool and models that default to disease-absent predictions benefit from this class imbalance. The balanced construction of DDT removes this advantage at the pointwise level, yet $\mathrm{PDRA}_{\text{DDT}}$ drops sharply: from 0.68 to 0.37 for GPT-4o and from 0.58 to 0.20 for Llama3.3-70B. Models that appear competent on individual cases cannot reliably distinguish both members of an adversarial pair, indicating fragile rather than systematic reasoning.

\textbf{Diagnostic convergence failure reflects systematic underdiagnosis.} DCT accuracy remains below 0.53 for all models and falls below 0.35 for several. Because DCT requires accepting the anchor diagnosis for two clinically dissimilar patients, success depends on recognizing atypical presentations. Healthy Recall exceeds Sick Recall by 30 to 50 absolute points, confirming a pervasive bias toward rejecting the anchor diagnosis when the clinical profile deviates from prototypical patterns.

\textbf{Medical pretraining improves pairwise consistency.} SRS reshuffles model rankings relative to pointwise accuracy. Both medically pretrained models achieve higher SRS than general-purpose models of comparable or larger scale, and exhibit smaller Sick--Healthy Recall gaps. HuatuoGPT-o1-8B attains the highest SRS overall without leading on any individual task. Nevertheless, the best SRS of 0.29 leaves substantial room for improvement. We analyze failure patterns in subsequent sections.

\subsection{Failure Mode Analysis}
\label{subsec:error-analysis}

\begin{figure}[htbp]
     \centering
     \includegraphics[width=0.48\textwidth]{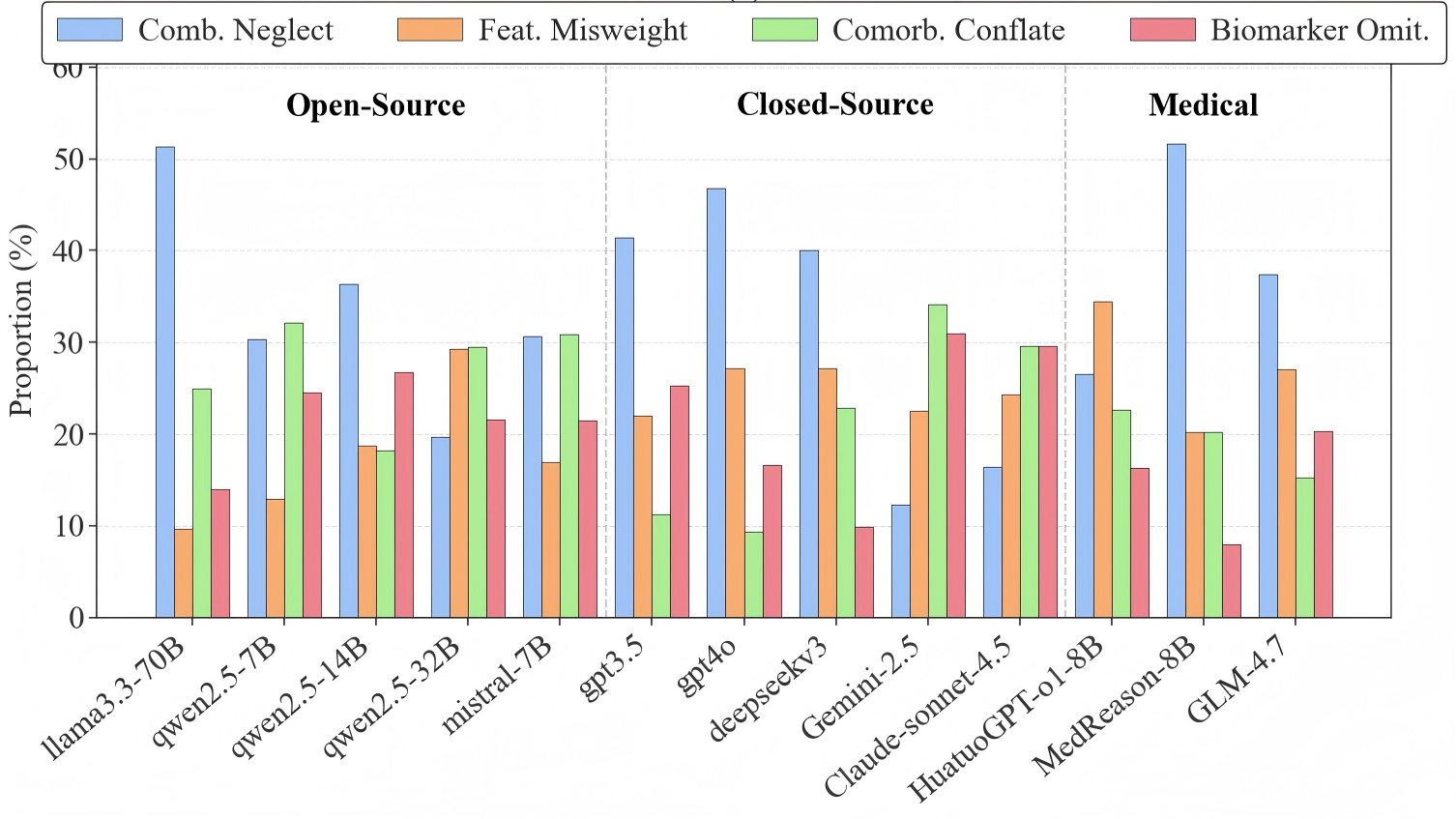}
     \caption{Distribution of four failure modes across 13 models.
     (a)~Overall proportions.
     (b)~Per-model breakdown grouped by model category.}
     \label{fig:wrong}
\end{figure}

To understand why models fail on adversarial pairs, we classify all 128,904 incorrect predictions into four clinically grounded failure modes through structured rationale analysis.

\textbf{Combinatorial Neglect} (34.0\%) is the dominant mode: models attend to individual clinical cues yet fail to integrate jointly decisive evidence. This error is particularly damaging in DDT, where distinguishing confusable patients requires synthesizing multiple discriminating features simultaneously.\textbf{Feature Misweighting} (23.1\%) reflects miscalibrated importance assigned to clinical indicators.\textbf{Comorbidity Conflation} (22.7\%) captures the inability to disentangle a primary diagnosis from coexisting chronic conditions. This mode is particularly relevant to DCT, where two patients sharing the same diagnosis present with divergent comorbidity profiles.\textbf{Biomarker Omission} (20.3\%) occurs when models reference key biomarkers in their rationales but neglect them in final decisions.

The per-model error profiles in Figure~\ref{fig:wrong} trace a developmental trajectory. At the weaker end of the spectrum, models such as Mistral-7B and Qwen2.5-7B allocate over 20\% of errors to Biomarker Omission and exhibit elevated Comorbidity Conflation, reflecting incomplete utilization of basic clinical indicators. Stronger general models largely resolve these shortcomings but converge on a new bottleneck: Combinatorial Neglect accounts for nearly half of all errors in DeepSeek-V3 and GPT-4o, revealing that multi-evidence integration remains fragile even at frontier scale. Domain-specialized models achieve the lowest Biomarker Omission rates yet simultaneously the highest Combinatorial Neglect, confirming that biomarker recognition and evidence integration are separable competencies. Meanwhile, Feature Misweighting persists uniformly across all categories, pointing to a structural limitation that neither scaling nor domain adaptation currently addresses. Representative cases appear in Appendix Table~\ref{tab:error}.

\subsection{Degradation Analysis}
\label{subsec:degradation}

\begin{figure*}[t]
    \centering
    \includegraphics[width=\textwidth]{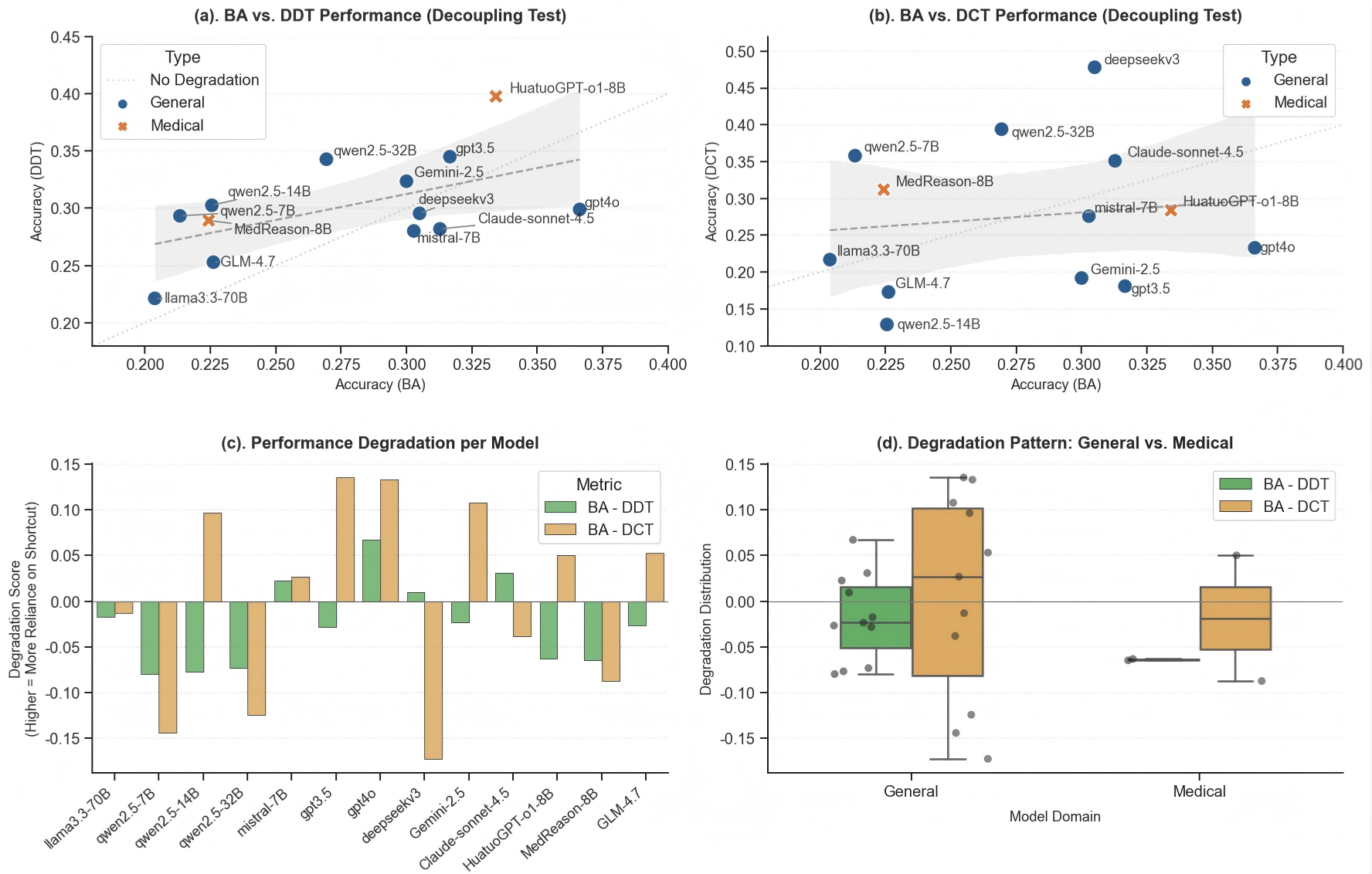}
    \caption{Performance degradation under adversarial evaluation.
    (a,b)~Correlation between BA accuracy and task accuracy for DDT and DCT respectively.
    (c)~Per-model degradation scores; positive values indicate performance loss.
    (d)~Distribution of degradation scores for general-purpose versus medical models.
    Blue circles denote general-purpose LLMs; orange crosses denote medical LLMs; the grey dashed line marks zero degradation.}
    \label{fig:degradation}
\end{figure*}

Figure~\ref{fig:degradation} shows that DDT and DCT degradation scores are nearly uncorrelated across models, exposing an orthogonality that pointwise accuracy alone conceals. A model that preserves performance under diagnostic divergence gains no corresponding advantage under diagnostic convergence. DeepSeek-V3 illustrates the extreme case: its DCT accuracy exceeds its own baseline while DDT degrades substantially, as though convergence and divergence reasoning draw on non-overlapping resources. HuatuoGPT-o1-8B, by contrast, maintains near-zero degradation on both tasks.

This dissociation maps cleanly onto training regime. Medical models cluster tightly at the zero-degradation origin regardless of which task is measured, achieving a task-general robustness that no individual general-purpose model replicates. General models instead distribute widely along one axis or the other, each brittle in its own idiosyncratic direction. Whether parameter scaling can substitute for this domain-acquired uniformity is the subject of Section~\ref{subsec:scaling}.

\subsection{Scaling Analysis}
\label{subsec:scaling}

\begin{figure}[t]
    \centering
    \includegraphics[width=\linewidth]{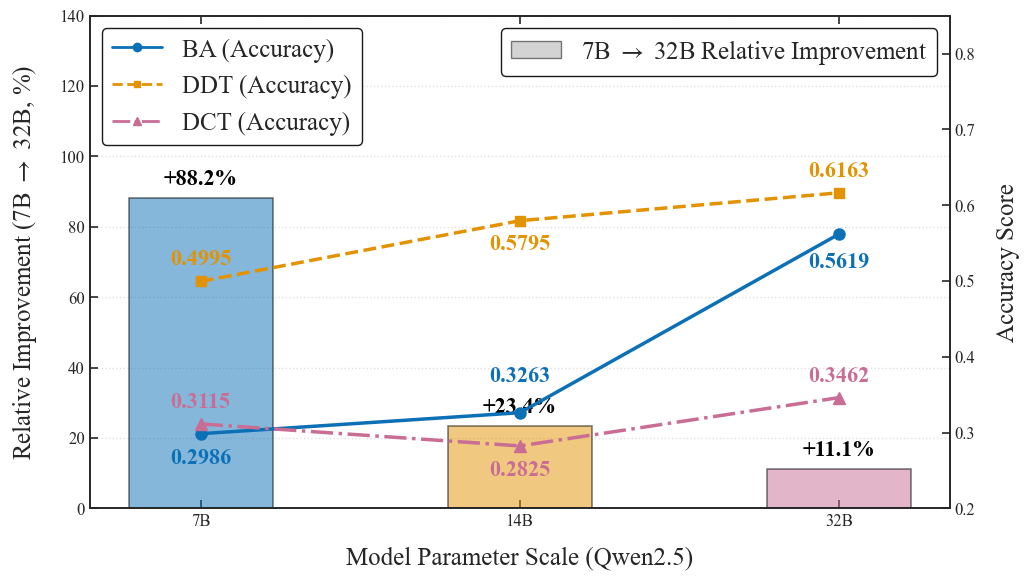}
    \caption{The Qwen2.5 series demonstrates scalability under BA, DDT, and DCT algorithms; line graphs show accuracy at 7B, 14B, and 32B resolutions; bar charts show the relative performance improvement from 7B to 32B resolution.}
    \label{fig:scaling}
\end{figure}

To isolate the effect of model scale from architecture and training data differences, we compare three Qwen2.5 variants that share the same design but differ in parameter count. As shown in Figure~\ref{fig:scaling}, scaling from 7B to 32B yields highly non-uniform returns across tasks. BA benefits most, rising from 0.30 to 0.56, an 88\% relative gain that reflects improved pattern recognition over the majority class. DDT improves moderately from 0.50 to 0.62, consistent with better surface-level feature matching. DCT, however, improves only marginally from 0.31 to 0.35, with a transient dip at 14B, indicating that the reasoning required to recognize atypical presentations resists parameter scaling almost entirely.

This asymmetry answers the question raised by Section~\ref{subsec:degradation}: scaling amplifies the very capabilities that BA already measures while leaving the adversarial robustness gap largely intact. The widening ratio between BA and DCT gains suggests that additional parameters preferentially consolidate prototypical pattern associations rather than cultivating the integrative reasoning that adversarial evaluation demands.

\subsection{Disease Category Stratification}
\label{subsec:stratification}

We map all 200 benchmark diseases to 17 ICD-based categories and test whether the degradation patterns identified above are clinically uniform. One-way ANOVA rejects uniformity for all three tasks ($F \geq 5.81$, $p < 0.001$ in each case).
Figure~\ref{fig:stratification} reveals a clear organizing principle. Categories defined by distinctive biomarker signatures within a single organ system, achieve stable accuracy across BA, DDT, and DCT, reflecting diagnoses that can be resolved through isolated indicator recognition. Categories requiring cross-system evidence integration, including metabolic, respiratory, and neoplasm-related diseases, show selective DCT collapse with statistical significance ($p < 0.001$ for metabolic and respiratory categories). This is precisely the setting where Combinatorial Neglect dominates (Section~\ref{subsec:error-analysis}). Notably, several categories exhibit DDT accuracy exceeding BA, confirming that the class-imbalance artifact identified in Section~\ref{subsec:main-results} operates at the category level as well. Therefore, adversarial fragility is not a uniform tax on performance but a domain-selective deficit that tracks the integrative complexity of the diagnostic task.

\section{Conclusion}

We introduce SUP-MIMIC, a multi-task clinical diagnosis benchmark that evaluates LLM robustness to contradictory evidence through adversarial patient pairs mined from MIMIC-IV ICU records. Our experiments reveal three consistent findings: pairwise evaluation exposes reasoning inconsistency that pointwise accuracy conceals, error profiles shift from basic omission toward integration failure as model capability increases, and parameter scaling preferentially improves baseline performance while leaving adversarial robustness largely unaddressed. Medical pretraining provides more uniform robustness across tasks but does not resolve the dominant bottleneck of multi-evidence integration. These results suggest that reliable clinical diagnosis requires targeted advances in combinatorial reasoning and atypical case recognition beyond what current scaling and domain adaptation strategies provide.

% This paper introduces SUP-MIMIC, a diagnostic reasoning benchmark constructed from MIMIC-IV-v3.1 for evaluating large language models under clinically challenging and adversarial settings. We design three complementary tasks to simulate difficult diagnostic scenarios involving conflicting clinical indicators, overlapping symptom patterns, and counterfactual perturbations. 

% Experimental results show that current LLMs still rely heavily on statistical correlation patterns rather than robust causal clinical reasoning. Although advanced models such as GPT-4o and HuatuoGPT-o1-8B demonstrate relatively strong performance on standard diagnostic tasks, their accuracy and stability decline substantially in adversarial settings involving contradictory indicators and atypical patient presentations. 

% Methodologically, our work is among the first to systematically incorporate adversarial indicator--outcome relationships into a large-scale medical benchmark for LLM evaluation. By explicitly constructing similar and dissimilar patient pairs with controlled feature perturbations, SUP-MIMIC provides a practical framework for analyzing clinical reasoning failures under distribution shifts and diagnostic conflicts. Our findings suggest several important directions for future research, including improving causal and counterfactual reasoning capabilities, enhancing atypical case recognition, and increasing diagnostic consistency and robustness for safety-critical medical applications.

\section*{Limitations}

First, to ensure diagnostic reliability and minimize the impact of outliers, we excluded diagnoses with fewer than 100 ICU admissions, which limited the coverage of rare diseases. However, physician examinations and final outcomes still demonstrate that the current benchmark has sufficient clinical validity and diagnostic representativeness. Future work may include manually annotated rare disease cases to mitigate selection bias and expand diagnostic coverage. Second, this study relied solely on structured EHR features for evaluation and did not incorporate multimodal information such as medical imaging and waveform data. Nevertheless, our feature set covers 448 clinical indicators, including multiple dimensions such as vital signs and laboratory tests, and has been validated by experts to possess sufficient diagnostic information. Future work could further integrate multimodal data to more comprehensively simulate real-world clinical decision-making scenarios.

\bibliography{custom}

\clearpage
%% 单栏（变双就给他删掉）
%% \onecolumn
\appendix

\section{Data Access and Licensing}

MIMIC-IV-v3.1 is a well-established, de-identified clinical database that has already undergone strict, professional anonymization according to HIPAA Safe Harbor standards. In this work, we utilize the MIMIC-IV-v3.1 dataset. Access to this database was approved after completion of the required safe research training (CITI program). The dataset is used in strict accordance with the PhysioNet Credentialed Health Data License. The artifacts and evaluation frameworks created in this study (SUP-MIMIC) will be released under the MIT License for non-commercial academic research purposes. During the preparation of this manuscript, AI assistants (such as ChatGPT/Claude) were utilized strictly for English language polishing, grammatical corrections, and text proofreading. The AI assistants were not involved in driving the core scientific conclusions, creating the dataset, or directly generating the primary experimental code.

\section{More experimental results}

\subsection{Feature Selection and Pair-Construction Framework}
\label{subsec:feature-selection}

To determine how many important features are sufficient for constructing reliable evaluation pairs, we vary $K \in \{5,10,\dots ,50\}$ and examine:
(i) how many distinct original variables are covered by the union of per-disease top-$K$ features (feature utilization), and
(ii) how LLM diagnostic accuracy and stability change across diseases.

For each $K$, we train a random forest on the full 448-dimensional space, extract the top-$K$ important features per disease, and take their union.
The size of this union is reported as \textit{Available Features} in Table~\ref{tab:20}, and its ratio to 448 as \textit{Feature Utilization}.
Thus, feature utilization measures how much of the original feature space is actually involved in constructing similar/dissimilar patient pairs, even though the LLM always receives all 448 features.

\begin{figure}[htbp]
     \centering
     \includegraphics[width=0.48\textwidth]{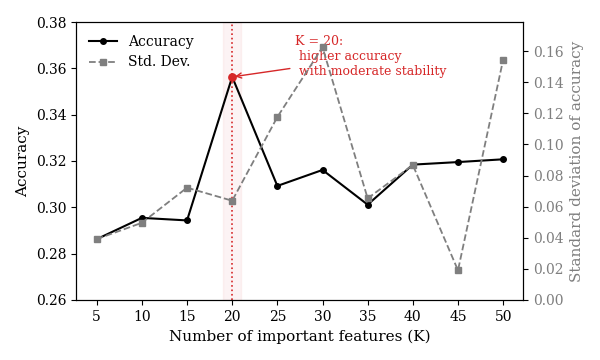}
     \caption{Trade-off between diagnostic accuracy and variance as a function of the number of important features (K), with K = 20 achieving the best balance.}
     \label{fig:20}
\end{figure}

As shown in Table~\ref{tab:20} and Figure~\ref{fig:20}, increasing $K$ from 5 to 20 raises feature utilization from 1.78\% (8/448) to 51.33\% (230/448), and accuracy from 0.2862 to a peak of 0.3563.
Small $K$ values therefore suffer from poor coverage: pairs are repeatedly formed based on a narrow subset of clinical attributes, underutilizing the diversity of the 448-dimensional space and yielding weaker performance across the 35 diseases.

When $K>20$, feature utilization continues to grow (57.36\%–64.51\%), but accuracy drops to about 0.31–0.32 and the variance across diseases increases markedly (e.g., standard deviation 0.0638 at $K=20$ vs 0.1177 at $K=25$ and 0.1626 at $K=30$).
This suggests that adding more “important” features beyond $K=20$ mainly introduces marginal or noisy variables, leading to more heterogeneous and less interpretable behavior.

Overall, $K=20$ strikes a good balance between coverage and stability.
It covers over half of the original variables, achieves the highest overall accuracy, and maintains moderate cross-disease variance.
We therefore adopt $K=20$ as the default for constructing similar and dissimilar patient pairs.
Importantly, this only affects pair construction: each patient is still represented by the full 448-dimensional feature vector as input to the LLM.

\begin{table}[htbp]
\centering
\begin{minipage}{\columnwidth}
\resizebox{\columnwidth}{!}{
\begin{tabular}{cccccc}
\hline
\textbf{Important} & \textbf{Available} & \textbf{Feature} & \textbf{Model} & \textbf{Accuracy} & \textbf{Std. Dev.} \\
\textbf{Features} & \textbf{Features} & \textbf{Utilization} & \textbf{Accuracy} & \textbf{Variance} & \textbf{(Accuracy)} \\
\hline
5   & 8   & 1.78\%  & 0.2862 & 0.00153  & 0.0391 \\
10  & 124 & 27.67\% & 0.2954 & 0.00246  & 0.0496 \\
15  & 150 & 33.48\% & 0.2943 & 0.00521  & 0.0722 \\
\textbf{20}  & \textbf{230} & \textbf{51.33\%} & \textbf{0.3563} & \textbf{0.00407} & \textbf{0.0638} \\
25  & 257 & 57.36\% & 0.3092 & 0.01384  & 0.1177 \\
30  & 261 & 58.25\% & 0.3161 & 0.02642  & 0.1626 \\
35  & 264 & 58.93\% & 0.3011 & 0.00426  & 0.0653 \\
40  & 268 & 59.82\% & 0.3184 & 0.00754  & 0.0868 \\
45  & 278 & 62.05\% & 0.3195 & 0.00036  & 0.0190 \\
50  & 289 & 64.51\% & 0.3207 & 0.02393  & 0.1546 \\
\hline
\end{tabular}
}
\caption{Comparison of model performance and feature utilization across different numbers of important features. Accuracy, variance, and standard deviation are reported for the overall (ALL) setting.}
\label{tab:20}
\end{minipage}
\end{table}

\subsection{Confusion Pattern \& Healthy Bias}

\begin{figure}[t]
    \centering
    \includegraphics[width=\linewidth]{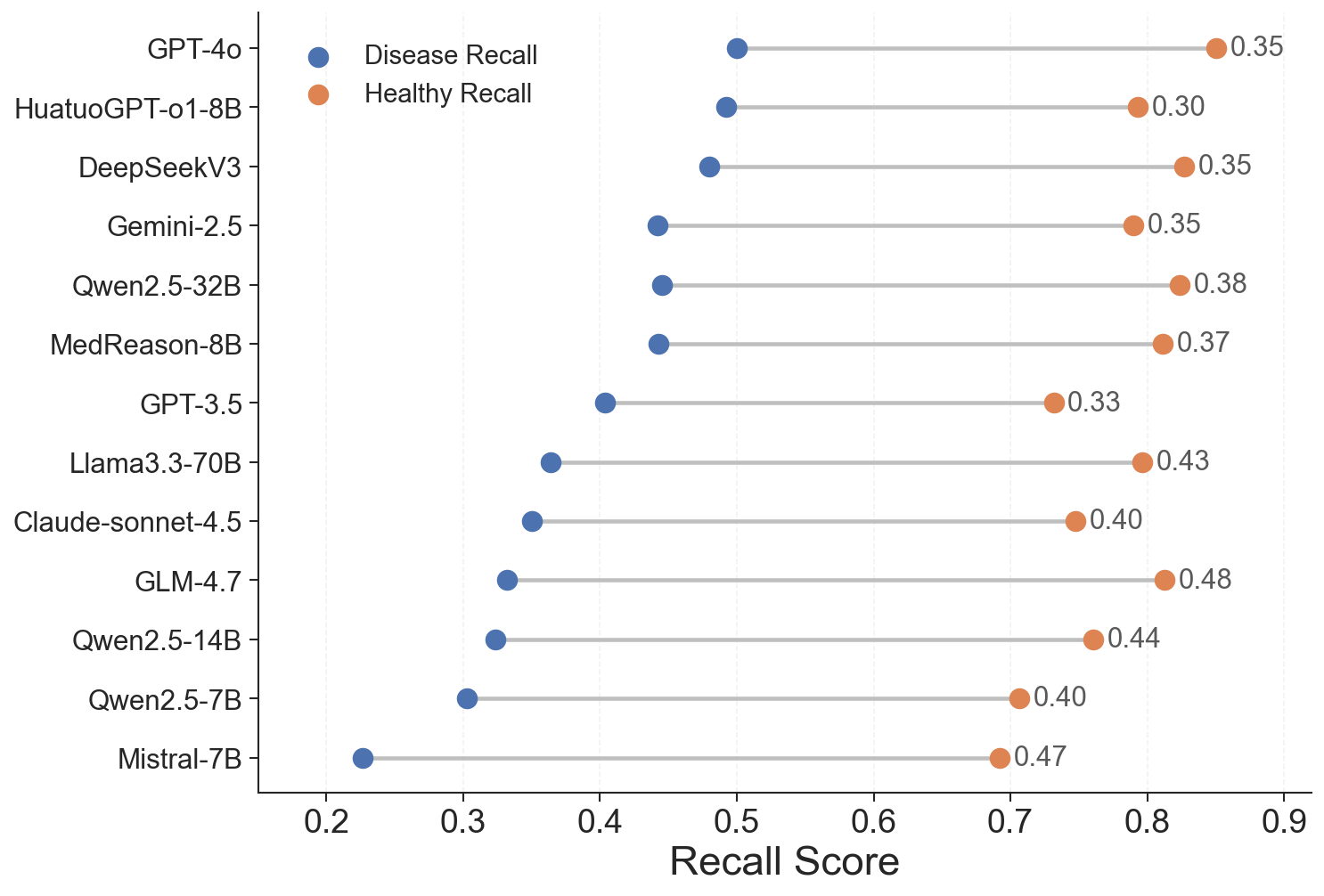}
    \caption{Diagnostic Performance Disparity Across Evaluated LLMs.
The dumbbell plot in it illustrates the performance trade-offs between Disease Recall (blue circles) and Healthy Recall  (orange circles). Models on the vertical axis are sorted in descending order of their Disease Recall proficiency. The horizontal grey lines represent the performance discrepancy within each model, with adjacent numerical annotations indicating the exact absolute gap.}
    \label{fig:exp4}
\end{figure}

Figure~\ref{fig:exp4} illustrates the trade-off between disease sensitivity and healthy-case recognition across different LLMs. 
A consistent discrepancy can be observed between positive-case recall and negative-case recall, where nearly all models achieve substantially higher recall on healthy samples than on disease samples. 
This phenomenon indicates a systematic healthy prediction bias, suggesting that current LLMs tend to favor conservative or non-pathological predictions under diagnostic uncertainty.

Among all evaluated models, GPT-4o and HuatuoGPT-o1 achieve the highest disease recall while maintaining relatively strong healthy-case recognition, demonstrating a more balanced diagnostic capability. 
In contrast, smaller open-source models exhibit both lower disease sensitivity and larger performance variance, indicating weaker robustness in challenging diagnostic scenarios.

\subsection{ROC/PR \& Threshold Sensitivity}
Finally, we analyze whether model confidence provides useful discrimination and how decision thresholds affect diagnostic risk. We construct ROC and PR curves using model confidence as the scoring function, then compute Equal Error Rate (EER) thresholds and Missed Diagnosis Rate (MDR) under both the default threshold and optimized thresholds.

AUC does not degrade from BA to CDT/DCT, with maximum $|\Delta\text{AUC}| < 0.025$. GPT-4o achieves AUC $\sim$0.83 uniformly across tasks, indicating that confidence retains some discriminative information even when default decisions are biased. At the default threshold of 0.5, all models exhibit healthy bias. EER thresholds cluster between 0.43 and 0.48, suggesting that lower thresholds are needed to balance recall. At EER thresholds, Sick Recall improves by 10--15 percentage points, for example from 52\% to 65\% for GPT-4o. MDR, our clinical risk metric, is reduced by 20--25\% at EER relative to the default threshold. Confidence gap analysis further shows that models assign systematically higher confidence to ``healthy'' predictions than to ``sick'' predictions.

\begin{figure}[h]
    \centering
    \includegraphics[width=\linewidth]{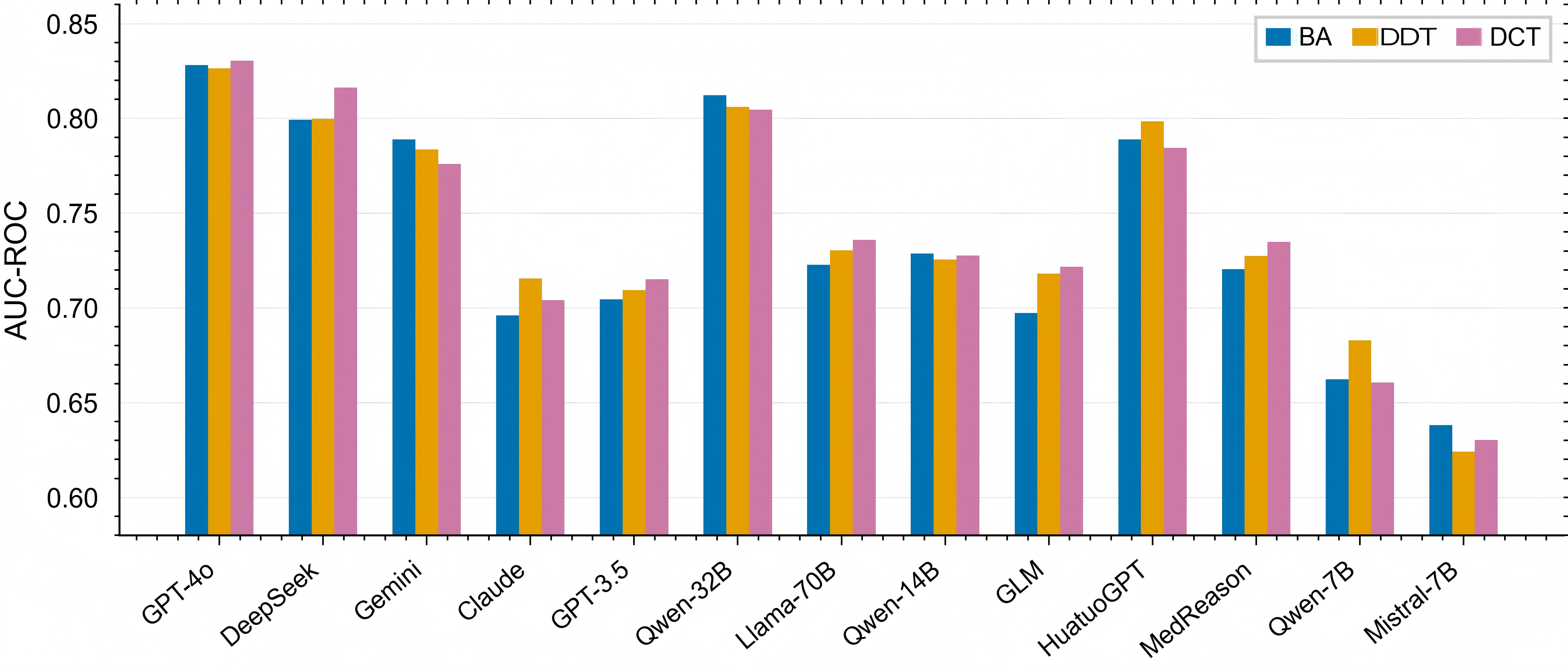}
    \caption{AUC-ROC by model and task type. Near-identical AUC across BA, CDT, and DCT confirms that confidence-based discrimination is task-independent.}
    \label{fig:exp8_roc}
\end{figure}

\begin{figure}[h]
    \centering
    \includegraphics[width=\linewidth]{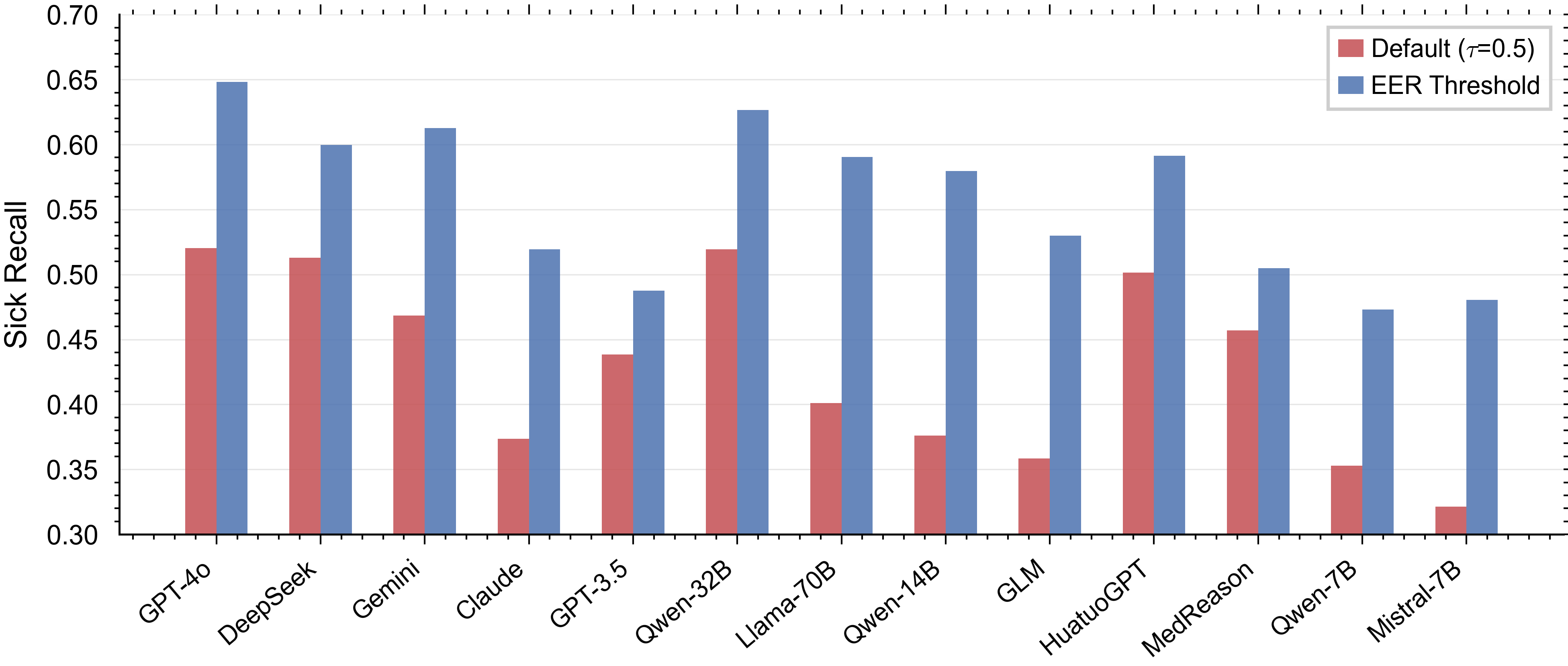}
    \caption{Sick Recall at default threshold ($\tau=0.5$) vs EER-optimized threshold. EER calibration improves Sick Recall by 10--15 percentage points across all models.}
    \label{fig:exp8_eer}
\end{figure}

\subsection{Benchmark Quality Assessment}

\label{subsec:quality}
We convened an expert panel comprising three practicing physicians to conduct a clinical quality assessment of the benchmark dataset. The panel consisted of one associate chief physician specializing in intensive care medicine from a Grade 3A (Class A Tertiary) teaching hospital, and two attending physicians from independent hospitals. All members hold post-graduate degrees and possess at least eight years of active clinical experience. As academic co-contributors to this study, they participated voluntarily for scientific collaboration purposes, and thus external commercial compensation was not applicable.

To evaluate feature efficacy, we randomly selected five diseases from the benchmark dataset and extracted the top 20 features identified for each disease via our Random Forest feature selection pipeline. The expert panel achieved a consensus that these features are tightly coupled with the core pathophysiological mechanisms of the target diagnoses. Consequently, these features were deemed highly appropriate for constructing clinically meaningful adversarial patient pairs.

We further assessed whether the 448-dimensional structured feature representation utilized in SUP-MIMIC conveys sufficient diagnostic information to sustain robust clinical reasoning. Operating independently and without access to free-text clinical notes or imaging reports, each physician made diagnostic determinations solely using the structured clinical indicators and evaluated which specific features effectively buttressed their decisions. The panel unanimously agreed that reliable diagnostics could be achieved based exclusively on the 448-dimensional structured features, subsequently stratifying them into four distinct categories. To quantify inter-annotator agreement during the feature efficacy evaluation, we computed Fleiss' $\kappa$ and pairwise Cohen's $\kappa$ coefficients. The overall Fleiss' $\kappa$ reached 0.9516, demonstrating near-perfect agreement among the annotators. The pairwise Cohen's $\kappa$ coefficients were 0.9492, 0.9456, and 0.9600, respectively. These statistical results rigorously substantiate that the feature selection pipeline employed in our benchmark construction exhibits exceptional clinical reliability, consistency, and high reproducibility.

\section{Prompts}

To illustrate the end-to-end prompting and reasoning process for a single intensive care unit (ICU) admission, we present an anonymized example of the actual request sent to the model. All patient features are directly derived from the publicly available MIMIC-IV-v3.1 critical care database. All personal identifiers have been removed, and numerical values are slightly perturbed to protect privacy while preserving clinical plausibility and relative magnitudes.

\begin{figure}[t]
\centering
\includegraphics[width=0.9\linewidth]{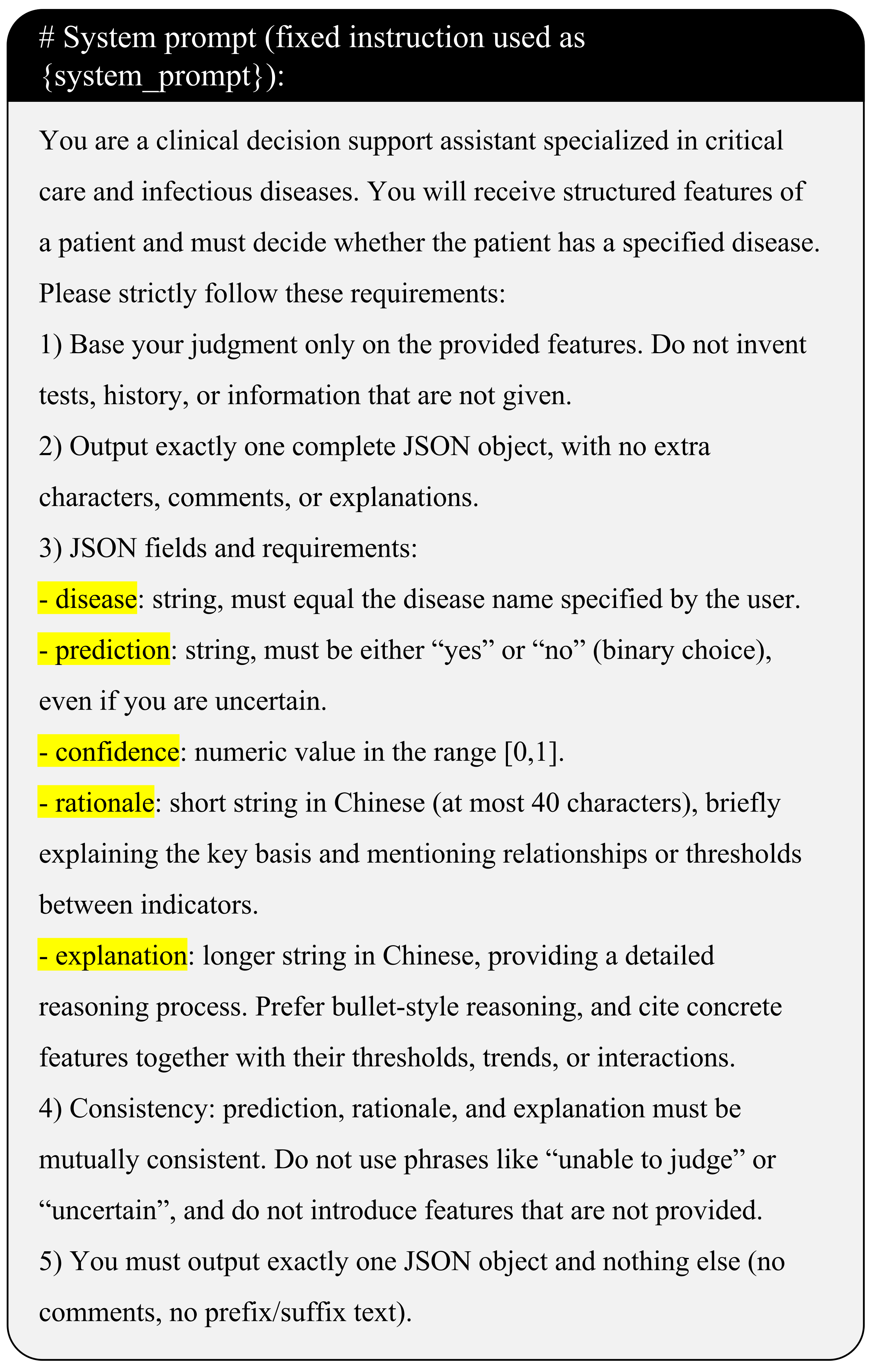}
\caption{System prompt used for clinical reasoning and diagnostic evaluation.}
\label{fig:system_prompt}
\end{figure}

\begin{figure}[b]
\centering
\includegraphics[width=0.9\linewidth]{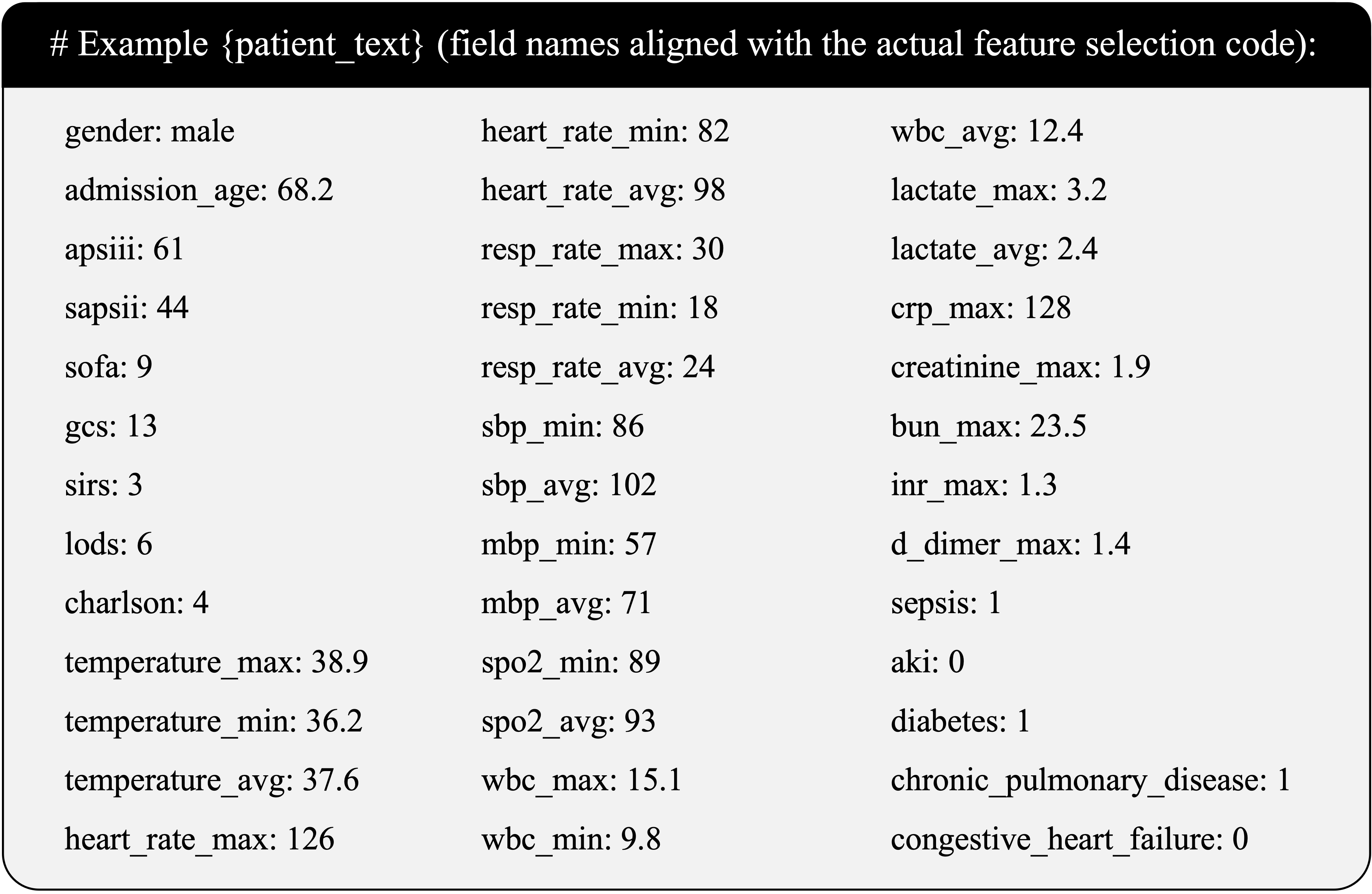}
\caption{Feature presentation format provided to the LLMs during inference.}
\label{fig:feature_display}
\end{figure}

Figures~\ref{fig:system_prompt} and~\ref{fig:feature_display} show representative request examples used in our experiments. Figure~\ref{fig:system_prompt} presents the system-level instruction used to standardize the reasoning behavior of the models, while Figure~\ref{fig:feature_display} illustrates how structured ICU features are converted into textual inputs for inference.

This block corresponds to the actual system messages used in our inference pipeline, where \texttt{\{patient\_text\}} is programmatically generated from tabular ICU features by the Python script described in the ``Labels of Features'' section. The generated text integrates heterogeneous clinical variables, including demographics, laboratory findings, vital signs, and so on into a unified natural-language representation for model reasoning.

\section{More error cases}

% 定义颜色命令（保持原配色不变）
\newcommand{\greentext}[1]{\textcolor[RGB]{65,117,5}{#1}}
\newcommand{\redtext}[1]{\textcolor[rgb]{0.82,0.01,0.11}{#1}}

\onecolumn
\scriptsize
\setlength{\tabcolsep}{2pt}   % 保持原列间距
\renewcommand{\arraystretch}{1.3} % 增加行高，提升可读性

\begin{longtable}{
>{\centering\arraybackslash}m{1.1cm}
>{\centering\arraybackslash}m{2.6cm}
>{\raggedright\arraybackslash\hspace{0pt}}m{9.0cm}
>{\centering\arraybackslash}m{2.1cm}
}

% 第一页完整表头（包含标题和标签）
\caption{Representative diagnostic error attribution cases from different LLMs. Green text indicates clinically reasonable and well-supported evidence; red text indicates misused, misinterpreted clinical evidence or flawed reasoning; highlighted background indicates missing evidence or unaddressed analytical gaps.}
\label{tab:error} \\
\toprule
\textbf{Case} &
\textbf{Disease} &
\textbf{Model Output Explanation} &
\textbf{Error Model} \\
\midrule
\endfirsthead

% 后续页面重复表头
\caption*{Table \ref{tab:error} (Continued)} \\
\toprule
\textbf{Case} &
\textbf{Disease} &
\textbf{Model Output Explanation} &
\textbf{Error Model} \\
\midrule
\endhead

% 所有页面底部显示的表尾
\midrule
\multicolumn{4}{r}{\textit{Continued on next page}} \\
\endfoot

% 最后一页底部显示的表尾
\bottomrule
\endlastfoot

% 表格内容（完全保留您的原文，仅修正语法错误）
DCT & 
Cerebrovascular Disease & 
The recorded variable \textit{\greentext{itself constitutes explicit positive evidence, indicating that the patient has cerebrovascular disease.}} In addition, the patient is of extremely advanced age (92 years) and has multiple comorbidities, including diabetes mellitus, congestive heart failure, renal disease, and sepsis, all of which are well-established risk factors for cerebrovascular events.\par On admission, the neurological status was profoundly impaired \redtext{(GCS = 3)}, suggesting a possible acute central nervous system event. Furthermore, the patient exhibited severe coagulation dysfunction \redtext{(INR up to 8.31, PT 92 s, platelet count 16–48 $\times 10^9$/L, fibrinogen approximately 90 mg/dL)}. In the context of \redtext{sepsis}, these abnormalities substantially increase the risk of intracranial hemorrhage or hemorrhagic transformation following ischemic stroke. At the same time, concurrent hypoxia and \redtext{shock} also raise the possibility of ischemic brain injury.\par Although imaging evidence is lacking, in a high-risk ICU setting, and to avoid underdiagnosis, the decision is ultimately made as "Yes," based on the confirmed positive history and multiple strongly supportive clinical indicators. &
claude-sonnet-4 \par(Error Type: Biomarker Omission) \\
\midrule

DCT & 
Cervical Spine Degeneration  & 
The patient is an \redtext{85-year-old} male. Advanced age is one of the most important risk factors for degenerative changes of the cervical spine, including intervertebral disc degeneration, uncovertebral or facet joint osteoarthritis, osteophyte formation, and spinal canal or neural foraminal stenosis. Epidemiological studies indicate that \redtext{the imaging detection rate of such changes in this age group is extremely high and nearly ubiquitous}.\par The patient is currently in a critical condition\redtext{ (SOFA score 20, GCS 3), complicated by sepsis and acute kidney injury (AKI)}. In this clinical context, \redtext{it is not feasible to reliably assess or exclude cervical degenerative changes based on symptoms and physical examination alone}. Following the principle of maximizing sensitivity and avoiding missed diagnoses, in the absence of definitive imaging evidence to the contrary, the presence of degenerative changes should be favored.\par In addition, the patient has comorbidities such as peripheral vascular disease, chronic pulmonary disease, and renal disease, which further suggest a high long-term risk of degenerative changes. There is no evidence supporting alternative specific etiologies such as inflammatory spinal disease or rheumatologic disorders. Although current laboratory abnormalities (infection, hepatic and renal dysfunction, and coagulation disturbances) explain the overall critical illness state, they do not directly conflict with or negate cervical spine degeneration.\par Therefore, despite the absence of direct imaging labels, given the extremely high age-related prior probability and the inability to adequately evaluate the condition in the current clinical setting. & 
deepseekv3 \par(Error Type: Feature Misweighting) \\
\midrule

DDT & 
Sleep Disorders & 
The patient is a 78-year-old male in a critical ICU/emergency condition, complicated by sepsis and acute kidney injury (AKI), with a high overall disease severity (APSIII 74, SAPSII 52, SOFA 4). Such a critically ill physiological state and environment are inherently prone to disturbances of the sleep--wake cycle and marked sleep fragmentation.\par 
In addition, the patient is of advanced age and has a history of dementia and cerebrovascular disease, all of which are important risk factors for sleep disorders, including insomnia, circadian rhythm disturbances, and sleep-related breathing disorders. A \textit{\greentext{GCS score of only 6}} indicates significantly impaired consciousness and strongly suggests the likelihood of ICU interventions such as sedation and mechanical ventilation. Together with the ICU environment---characterized by noise, frequent nursing interventions, and disrupted light--dark cycles---these factors further contribute to the disruption of normal sleep architecture.\par 
\colorbox{yellow!30}{(Though the model correctly identified related risk factors and impaired consciousness, }\par 
\colorbox{yellow!30}{it incorrectly conflated transient critical-care sleep disruption with a formal diagnosis of} \par
\colorbox{yellow!30}{chronic or clinically established sleep disorders.)} \par
The patient also exhibits pronounced systemic inflammation and physiological stress responses, including leukocytosis (peak WBC 18.3), fever (Tmax 38°C), tachycardia (peak HR 151), metabolic alkalosis, hypernatremia (maximum Na 150), and anemia (lowest Hb 7.5). All of these factors are closely associated with delirium and sleep disturbances. At present, there is no evidence that clearly rules out sleep disorders. In this high-risk critical care setting, to avoid underdiagnosis, the presence of sleep disorders is therefore favored.\par
&
Gemini-2.5 \par(Error Type: Feature Comorbidity Conflation) \\
\midrule

BA &
Coagulation Defect &
The patient is in a state of severe sepsis/shock (sepsis = 1, SOFA = 15, peak lactate 13.8 mmol/L, AKI = 1), under which acquired coagulopathy is highly likely, including disseminated intravascular coagulation (DIC) or liver-related coagulation abnormalities. Multiple coagulation and platelet parameters show clear abnormalities, providing strong evidence of a coagulation defect:\par 1)Marked thrombocytopenia: \textit{\greentext{Platelet count on ICU admission was 61 × 10$^9$/L, with a nadir of 34 × 10$^9$/L, consistent with moderate-to-severe thrombocytopenia}}.\par 2)Significant coagulation dysfunction: \textit{\greentext{PT up to 26.6 s, INR up to 2.6}}, and \textit{\greentext{PTT up to 52.5 s}}, all indicating prolonged coagulation pathways.\par 3)Reduced fibrinogen: \textit{\greentext{Fibrinogen decreased to 132–155 mg/dL}}, suggesting consumptive hypofibrinogenemia.\par 4)Severe hepatic dysfunction: \textit{\greentext{AST peaked at 3404 U/L, ALT at 357 U/L, total bilirubin up to 23.7 mg/dL, with a MELD score of 25, indicating impaired synthesis of coagulation factors}}.\par 5)Severe anemia: \textit{\greentext{Hemoglobin decreased to a nadir of 4.8 g/dL}}, suggesting possible bleeding or consumption of clotting factors.\par
\colorbox{yellow!30}{It overemphasizes isolated abnormal coagulation markers without jointly considering}\par
\colorbox{yellow!30}{the dynamic recovery trend and the absence of definitive evidence for active DIC or }\par
\colorbox{yellow!30}{persistent coagulation failure.} \par
Although partial recovery of some parameters was observed after treatment, including improvement in fibrinogen and platelet count (Fbg 477 mg/dL, PLT 496 × 10$^9$/L) and a reduction of INR to 1.4, a prior classic triad of "thrombocytopenia + prolonged PT/INR/PTT + hypofibrinogenemia" was clearly present. In the context of sepsis, this constellation is highly consistent with DIC or acquired coagulation defect. An isolated low INR value at certain time points (0.7) is insufficient to negate previously sustained abnormalities.\par Taken together, given the high-risk clinical context and the strong supporting evidence above, the patient is judged to have a coagulation defect. &
GLM-4.7 \par(Error Type: Combinatorial Neglect) \\
\midrule

DDT & 
Altered Mental Status & 
A comprehensive assessment of whether the patient has altered mental status is as follows:\par 1)Direct evidence of consciousness level: The most direct indicator of current consciousness is a GCS score of 15, suggesting that the patient is awake, alert, and generally oriented. This represents strong negative evidence for altered mental status. In critically ill settings, significant mental status changes (e.g., delirium, lethargy, confusion, or disorientation) are typically accompanied by a decreased GCS score or abnormal sedation-related assessments; however, no such findings are present here.\par 2)Presence of risk factors for altered mental status without corresponding objective findings: The patient has \textit{\greentext{sepsis (SIRS = 3, WBC elevated to 13.9)}}, \textit{\greentext{transient hypotension (minimum SBP 73.5)}}, \textit{\greentext{marked hypoxemia (PaO2/FiO2 as low as 74, SpO2 down to 82\%)}}, acid–base fluctuations (pH 7.28–7.54), and\textit{\greentext{ transient lactate elevation (peak 4.5)}}. All of these factors may increase the risk of delirium or metabolic encephalopathy. However, there is currently no objective evidence of corresponding impairment in consciousness.\par 3)Metabolic and electrolyte abnormalities are insufficient to support definite encephalopathy: Sodium nadir is 131, indicating only mild hyponatremia. Glucose ranges from 77 to 195, without significant hypoglycemia or hyperosmolar state. PaCO2 peak is 48, not consistent with significant hypercapnia. BUN peaks at 32, providing insufficient evidence for uremic encephalopathy. Total bilirubin peaks at 1.6 and MELD is approximately 17.5, which does not support definite hepatic encephalopathy. Variations in calcium and magnesium are also insufficient to independently explain significant cognitive impairment.\par 4)Overall assessment: Although the patient has multiple high-risk factors for delirium or encephalopathy, \redtext{the most robust and direct evidence of mental status (GCS = 15)} remains dominant. There are also no clear positive psychiatric or objective neurological findings indicating altered consciousness. Therefore, the current assessment leans toward "no clear evidence of altered mental status." However, given the critical illness context, continued monitoring for potential delirium or encephalopathy is still warranted. & 
GPT-3.5 \par(Error Type: Comorbidity Conflation) \\
\midrule

BA & 
Arterial Embolism and Thrombosis & 
The patient was in a state of severe \redtext{sepsis/septic shock (SOFA 17, SIRS 4)}, with extremely severe metabolic acidosis and\redtext{ hyperlactatemia} occurring early in the course (peak lactate 18.2 mmol/L, lowest pH 7.14), accompanied by \redtext{acute kidney injury} (initial Cr 4.3, subsequently improved) and marked \redtext{"shock liver"} manifestations (peak AST/ALT 4296/2270), suggesting severe systemic hypoperfusion and tissue hypoxia.\par At the same time, there is clear evidence of \redtext{DIC/consumptive coagulopathy}: \redtext{INR peaked at 4.3; PT/PTT} were significantly prolonged; fibrinogen decreased to 117–187 mg/dL; and platelet count dropped to a nadir of 64 × 10$^9$/L.\par Severe sepsis complicated by DIC commonly leads to arterial and microcirculatory thrombosis, which can subsequently cause ischemic injury to organs such as the liver and kidneys. The current laboratory abnormalities and organ dysfunction are highly consistent with this pathophysiological process.\par Although direct imaging evidence is lacking, in a high-risk ICU setting, the condition should be managed as positive in order to avoid missed diagnosis. In addition, diagnosis\_long\_title1 = 418 is highly likely to correspond to "arterial embolism and thrombosis," providing further support. At present, there is no clear evidence against the diagnosis (such as normal coagulation function or imaging definitively excluding thrombosis).\par Overall assessment: the findings support the presence of arterial embolism and thrombosis. &
GPT-4o \par(Error Type: Comorbidity Conflation) \\
\midrule

DCT &
Coagulation Defect &
Based on a standardized clinical differential diagnostic analysis across multiple disease categories: this patient is a critically ill ICU patient with concomitant \textit{\redtext{sepsis (sepsis = 1.0) and malignant cancer (malignant\_cancer = 1.0), both of which are high-risk predisposing factors for coagulation disorders/DIC}}. Multiple coagulation-related tests show clear abnormalities:\par 1)\textit{\greentext{Marked thrombocytopenia with substantial fluctuation}}: the initial platelet count was 100 × 10$^9$/L, decreased to a nadir of only 5 × 10$^9$/L, and remained at 29 × 10$^9$/L on the last measurement, suggesting severe consumptive thrombocytopenia;\par 2)\textit{\greentext{Prolonged coagulation times}}: PT increased to 15.5 s, INR increased to 1.4, and PTT reached 44.1 s, indicating impairment of the coagulation cascade;\par 3)Early \textit{\greentext{significant reduction in fibrinogen}}: the lowest fibrinogen level was 126 mg/dL (below the normal range of 200–400 mg/dL), consistent with a consumptive coagulation process. Although fibrinogen later increased to >400 mg/dL, this may be related to an acute-phase response or replacement therapy and does not negate the prior existence of a coagulation defect;\par 4)\textit{\greentext{Marked decreases in Hb/Hct}}: hemoglobin dropped to a nadir of 5.4 g/dL and hematocrit to 18\%, which, in the setting of severe thrombocytopenia, suggests potential bleeding or ongoing consumption.\par In addition, liver function and MELD score (8 points) do not indicate severe liver failure and therefore do not support the alternative explanation of "coagulation factor deficiency caused solely by hepatic failure."\par Taken together, considering the high-risk underlying diseases and multiple abnormalities in coagulation parameters, the overall presentation is highly consistent with a "coagulation defect/coagulopathy."  & 
HuatuoGPT-o1-8B \par(Error Type: Combinatorial Neglect) \\
\midrule

DCT & 
Malignant Neoplasm of the Trachea and Lung & 
The patient is an \redtext{elderly male} \redtext{smoker} with \redtext{concomitant chronic pulmonary disease (COPD)}, placing him in a high-risk population for lung cancer. The medical history already documents the presence of malignant cancer \redtext{(malignant\_cancer = 1)}, while no distant metastasis is recorded (metastatic\_solid\_tumor = 0), which is highly consistent with the clinical background of a primary pulmonary malignancy.\par On admission, the patient presented with tachypnea, \redtext{hypoxemia} (PaO2/FiO2 approximately 231–250), and a markedly elevated A–a gradient, suggesting the presence of substantial pulmonary pathology. At the same time, the patient had \redtext{sepsis and significant neutrophilia}, which clinically fits the possibility of \redtext{"post-obstructive infection caused by an airway mass."} Although direct imaging evidence is currently lacking, the available data also do not provide clear evidence against pulmonary/tracheal malignancy (such as negative imaging findings or definite evidence of a primary tumor at another site).\par Considering the high-risk background, abnormal respiratory physiology, and infectious manifestations, and following the principle of maximizing sensitivity and avoiding missed diagnosis, the presence of malignant neoplasm of the trachea and lung should be highly suspected; therefore, the assessment is inclined toward "Yes." & 
llama3.3-70B \par(Error Type: Biomarker Omission) \\
\midrule

DCT &
Alcoholic Psychosis & 
Based on a standardized clinical differential diagnostic analysis across multiple disease categories: there is \redtext{no documented history of heavy alcohol consumption, recent alcohol cessation, or withdrawal-related events, nor are there core symptoms} suggestive of alcohol withdrawal or alcoholic psychosis, such as hallucinations, marked agitation, tremor, or seizures. The degree of impaired consciousness is limited to a GCS score of 13, and in the context of the overall clinical condition, infectious and metabolic factors represent \redtext{a more reasonable alternative explanation}.\par Specifically, the patient has clear sepsis (SIRS = 3, WBC $\geq$ 17 × 10$^9$/L, with marked neutrophilia), hypoxemia (lowest P/F ratio 158), electrolyte disturbance (hyponatremia 126–131 mmol/L), and acid–base fluctuations (pH 7.24–7.54), all of which can lead to\redtext{ infection-related delirium or metabolic encephalopathy}.\par Laboratory testing also does not support chronic alcohol-related injury: AST/ALT, bilirubin, and INR are essentially normal; albumin is near normal; and platelet count is not decreased but instead shows reactive elevation (up to 1062 × 10$^9$/L), which is\redtext{ inconsistent with alcohol-related bone marrow suppression}. In addition, there are no signs suggestive of Wernicke encephalopathy (such as ophthalmoplegia or ataxia) or other specific alcohol-related neurological manifestations.\par Overall, the current evidence is insufficient to support a diagnosis of alcoholic psychosis, while there is relatively clear negative evidence and stronger alternative etiologies. Therefore, the final assessment is "No." & 
MedReason-8B \par(Error Type: Comorbidity Conflation) \\
\midrule

BA & 
Pneumothorax & 
Risk and background: The patient is elderly, a smoker, and has chronic lung disease, which confers a risk for secondary spontaneous pneumothorax; however, risk factors alone are insufficient for diagnosis.\par Initial presentation: On admission, the patient had severe hypoxemia (SpO2 90\%, PaO2/FiO2 73.75, extremely elevated A–a gradient of 428.9), accompanied by marked respiratory acidosis and hypercapnia (pH 7.28, \redtext{PaCO2 66, later rising to 120}), together with tachycardia and increased respiratory rate.\par Pathophysiological inference: Pneumothorax (especially tension pneumothorax) typically presents with sudden dyspnea and hypoxemia, while PaCO2 is often normal or decreased early due to hyperventilation. If severe hypercapnia occurs, it is usually accompanied by significant circulatory compromise (hypotension and elevated lactate). In this case, hemodynamics were relatively stable (mean arterial pressure 56–73, without shock), and lactate remained persistently low (0.5–0.8), which is inconsistent with the characteristics of tension pneumothorax.\par Evidence supporting alternative diagnoses: Marked hypercapnia and elevated base excess (BE up to 9) suggest \redtext{a pattern of chronic CO2 retention/acute exacerbation of COPD}. At the same time, significant neutrophilia (WBC up to 23.1, NE\% up to 93.2) supports infection or pulmonary inflammation triggering AECOPD/pneumonia-related ventilation–perfusion impairment.\par Clinical course changes: Oxygenation improved markedly within a short period (PaO2/FiO2 increased to 287, A–a gradient decreased to 65.78, SpO2 increased to 99\%), which is more consistent with reversible hypoxemia and hypoventilation corrected by ventilation/oxygen therapy, rather than persistent pneumothorax without drainage.\par Key imaging evidence: No imaging labels of 1 or -1 providing positive or highly suspicious evidence were available (treated as unknown); therefore, there is no direct evidence supporting pneumothorax.\par Overall assessment: In the absence of positive imaging findings or highly suspicious signs, the current physiological features and disease course are more consistent with hypoxemia and hypercapnia caused by \redtext{AECOPD/infection}, while lacking the hemodynamic and metabolic evidence characteristic of tension pneumothorax. Pneumothorax is therefore considered unlikely. & 
mistral-7B \par(Error Type: Feature Misweighting) \\
\midrule

DDT & 
Disorders of Mineral Metabolism & 
There is clear evidence of mineral metabolism disturbance:\par1)\textit{\greentext{Hypocalcemia}}: Total calcium decreased to a nadir of 7.0 mg/dL; after correction for albumin 3.0 g/dL, the corrected calcium remained low at approximately 7.8 mg/dL. \textit{\redtext{Ionized calcium reached a nadirof 1.07–1.09 mmol/L (reference range 1.12–1.32)}}, confirming true hypocalcemia. This occurred together with marked metabolic alkalosis (maximum pH 7.51, HCO3-/TCO2 up to 42, BE up to +14), and alkalosis further decreases ionized calcium, consistent with the physiological mechanism of hypocalcemia.\par2) Phosphate abnormalities: Serum phosphate peaked at 6.6 mg/dL, representing significant hyperphosphatemia, while\textit{\redtext{hypophosphatemia (1.9 mg/dL) also occurred}}, indicating \textit{\greentext{fluctuating phosphate metabolism imbalance}}.\par3) Magnesium fluctuations: Magnesium ranged from 1.8–2.5 mg/dL. Although mostly within or near the normal range, there was a tendency toward mild hypermagnesemia, consistent with changes in renal function.\par4) Concurrent AKI and sepsis: The patient had AKI (Cr increased to 3.1 mg/dL, BUN to 37) together with sepsis, both of which commonly lead to hypocalcemia and hyperphosphatemia (through reduced renal excretion, calcium-phosphate deposition, inflammation, and disturbances of the PTH/vitamin D axis).\par5) Hypoalbuminemia may underestimate total calcium, but low ionized calcium directly excludes the negative explanation of "pseudohypocalcemia."\par In summary, the patient has pathophysiologically consistent hypocalcemia and hyperphosphatemia (with fluctuating hypophosphatemia), together with mild magnesium abnormalities, clearly meeting the criteria for a disorder of mineral metabolism. & 
qwen2.5-7B \par(Error Type: Biomarker Omission) \\
\midrule

DCT & 
Adrenal Disorders & 
On admission, the patient presented with \textit{\greentext{shock and severe lactic acidosis}} (SBP 84 mmHg, lactate 14.2, pH 7.03), \textit{\greentext{accompanied by severe hyponatremia}} (lowest Na 111). This phenotype can resemble an adrenal crisis. However, at the same time, the \redtext{potassium level was profoundly low} (initial K 1.0 mmol/L), which is inconsistent with the typical hyperkalemia seen in primary adrenal insufficiency and therefore represents strong negative evidence against this diagnosis.\par The subsequent transient severe hyperkalemia (peak 7.9) is consistent with extreme \redtext{rhabdomyolysis} (peak CK 183645), acidosis, and AKI, all of which \redtext{can independently explain the hyperkalemia} without requiring adrenal insufficiency as an explanation. Throughout the course, there was no hypoglycemia (lowest glucose 71), and no eosinophilia initially (initial value 0\%); the mild eosinophilia observed later lacks specificity.\par There were also no key features suggestive of hyperfunctioning adrenal disease: no sustained or paroxysmal hypertension (in contrast, hypotension predominated), no hypokalemic metabolic alkalosis suggestive of primary aldosteronism, and no typical manifestations of pheochromocytoma crisis.\par Although there was transient coagulation abnormality/DIC risk, which theoretically could cause adrenal hemorrhage, the patient's blood pressure and electrolyte abnormalities were rapidly corrected, and there was no persistent refractory shock or sustained "hyponatremia + hyperkalemia" combination supporting adrenal crisis.\par Overall, the current evidence is more consistent with severe shock complicated by rhabdomyolysis and fluid/metabolic disturbances causing the electrolyte abnormalities, rather than adrenal disorders. If refractory hypotension or recurrent hyponatremia develops subsequently, reassessment with random cortisol/ACTH testing and, if necessary, a short stimulation test is recommended to exclude CIRCI. & 
qwen2.5-14B \par(Error Type: Biomarker Omission) \\
\midrule

DCT & 
Drug Abuse & 
The patient is a 22-year-old male with no significant chronic medical history. On ICU admission, a GCS score of 7 indicated acute suppression of consciousness. Arterial blood gas analysis showed \textit{\greentext{pCO2 as high as 74 mmHg}}, consistent with a \textit{\greentext{respiratory depression phenotype}} (commonly seen in opioid or sedative overdose).\par At the same time, there was \textit{\redtext{extremely severe rhabdomyolysis and acute kidney injury: peak CK >120,000, LDH >15,000, potassium 8.0 mmol/L, phosphate 8.1 mg/dL, hypocalcemia, and creatinine peaking at 11.5 mg/dL.}} These findings are consistent with AKI secondary to rhabdomyolysis caused by prolonged coma/prolonged immobilization with compression or toxin/drug-induced rhabdomyolysis.\par Liver enzymes showed \textit{\greentext{explosive elevation}} (AST 16074, ALT 6549) followed by rapid decline, with only mild bilirubin elevation, a pattern typical of ischemic/toxic liver injury (commonly seen in overdose or post-shock liver injury) and inconsistent with chronic liver disease.\par On admission, the patient also exhibited stress hyperglycemia (424 mg/dL), \textit{\greentext{marked fluctuations in blood pressure and heart rate}} (SBP up to 223, heart rate up to 138), consistent with stimulant/catecholamine storm or severe stress response.\par Although "sepsis" was documented, lactate remained repeatedly not elevated ($\leq$ 2.0), and while WBC was initially elevated, it rapidly declined thereafter; evidence supporting infection is therefore weak, and the SIRS presentation may instead have been caused by rhabdomyolysis/toxic injury.\par In a young patient without a history of statin use, myopathy, or trauma, the most reasonable unifying causal chain is drug/toxin abuse or overdose leading to suppression of consciousness → prolonged compression/immobilization → rhabdomyolysis → hyperkalemia/AKI, together with ischemic/toxic liver injury. & 
qwen2.5-32B \par(Error Type: Feature Misweighting) \\

\end{longtable}

\normalsize

\twocolumn

\end{document}